\documentclass[11pt,a4paper]{article}

\usepackage[utf8]{inputenc}
\usepackage[T1]{fontenc}
\usepackage[english]{babel}
\usepackage{amsmath,amssymb,amsthm}
\usepackage{algorithm,algorithmic}
\usepackage{booktabs}
\usepackage{graphicx}
\usepackage{hyperref}
\usepackage{cleveref}
\usepackage{geometry}
\usepackage{multirow}
\usepackage{subcaption}
\usepackage{caption}
\usepackage{xcolor}
\usepackage{bm}
\usepackage{nicefrac}
\usepackage{microtype}
\usepackage{gensymb}

\hypersetup{colorlinks=true,linkcolor=blue!60!black,citecolor=green!50!black,urlcolor=blue!70!black}

\newcommand{\R}{\mathbb{R}}
\newcommand{\E}{\mathbb{E}}
\newcommand{\norm}[1]{\lVert #1 \rVert}

\newcommand{\mat}[1]{\bm{#1}}
\newcommand{\vect}[1]{\bm{#1}}
\newcommand{\Hess}{\mat{H}}

\newtheorem{definition}{Definition}

\title{Hessian Surgery: Class-Targeted Post-Hoc Rebalancing\\
via Hessian Spike Perturbation}

\author{
  \begin{tabular}{c@{\hspace{2cm}}c}
    Hugo Vigna & Samuel Bontemps \\
    \textit{CentraleSupélec -- Université Paris-Saclay} & \textit{ESILV -- Léonard de Vinci} \\
    \texttt{vignahugo@gmail.com} & \texttt{samuelbontemps2503@gmail.com}
  \end{tabular}
}

\date{\today}

\begin{document}
\maketitle

\begin{abstract}
The Hessian spectrum of trained deep networks exhibits a characteristic
structure: a continuous \emph{bulk} of near-zero eigenvalues and a small
number of large \emph{outlier eigenvalues} (spikes), confirming the
relevance of Random Matrix Theory in deep learning. The spike count
matches the number of classes minus one. While prior work has described
this structure, no method has exploited it \emph{operationally} to
improve classification performance.

We propose \textbf{Hessian Surgery}, a post-hoc optimization method
that directly perturbs model weights along spike eigenvectors to
rebalance per-class accuracy without retraining. We introduce
(i)~a \emph{spike--class sensitivity matrix} that quantifies the
directional derivative of each class's accuracy along each spike
eigenvector, (ii)~a constrained optimization of perturbation
coefficients that targets weak classes while preserving strong ones,
and (iii)~an adaptive amplitude control that raises or lowers the
perturbation budget based on iteration-level improvement signals.

On ResNet-50 trained on CIFAR-10, we reduce the standard deviation of
class accuracy by 36\% ($8.57\% \to 5.52\%$) with negligible impact
on global accuracy ($-$0.2\,pp), with the weakest class (cat)
gaining +7.5\,pp.
Hessian Surgery outperforms both focal loss and class-balanced
fine-tuning on class rebalancing, while operating entirely post-hoc.
The method requires only a mini-batch of a few hundred images for Hessian
estimation and a held-out evaluation set, and completes in minutes.
We further validate the structural role of spikes through bulk-walk
experiments showing that perturbations orthogonal to spikes leave
class-level performance invariant. On CIFAR-100, sequential deflated Surgery with 45 out of 99 spikes reduces $\sigma$
 by 1.6\% on the held-out set.
On ISIC-2019, a severely imbalanced 8-class medical imaging dataset,
Hessian Surgery alone marginally surpasses focal-loss fine-tuning in
rebalancing, and Hessian Surgery applied after Focal Loss fine-tuning reduces $\sigma$ by 30.5\% relative
($23.6\to16.4\%$) while modestly improving balanced accuracy over
Focal Loss fine-tuning alone ($51.5\%$ vs.\ $49.2\%$).
When applied on top of class-balanced fine-tuning - a stronger
training-time baseline that already reaches 57.7\% balanced accuracy - HS still reduces $\sigma$ by an additional $3.5$\,pp ($14.8\to11.3\%$),
without a balanced-accuracy gain. The \emph{effective rank} of the
spike--class sensitivity matrix contracts from $\approx 4$ on
cross-entropy-trained CIFAR-10 to $\approx 2$ on class-balanced ISIC-2019, which provides a candidate for quantitative explanation of the diminished gain on the latter. With only two regimes measured we cannot calibrate this as a numerical threshold; we report it as a
working hypothesis for which axes of redistribution remain
exploitable, to be tested on further architectures and datasets.
\end{abstract}


\section{Introduction}
\label{sec:intro}

Deep neural networks trained on classification tasks exhibit a
characteristic spectrum for the Hessian of the training objective
$\mat{H} = \nabla^2_\theta \mathcal{L}(\theta)$: a continuous
\emph{bulk} of eigenvalues clustered near zero, and $C{-}1$ large
isolated eigenvalues called \emph{spikes} or \emph{outliers}, where $C$ is the number of classes~\cite{sagun2017,papyan2020,ghorbani2019}.
The present work focuses on convolutional architectures for image classification (ResNet-50 on CIFAR-10/100 and ISIC-2019); the spike--bulk structure has been documented in this regime
specifically. This spectral decomposition has been extensively studied from a descriptive standpoint-connecting spikes to inter-class structure in the feature space, bulk eigenvalues to intra-class
variability, and the overall spectrum shape to generalization
properties.

Existing work uses this structure descriptively: the spectrum is
observed and characterized but not used to act on the trained model.
If each spike encodes a specific inter-class structure, then weight
perturbations along these directions should produce class-specific
effects that can be optimized. We use this property to define
\textbf{Hessian Surgery}, a method that:
\begin{enumerate}
    \item Measures the sensitivity of each class's accuracy
      to perturbations along each spike eigenvector, producing a
      \emph{spike--class sensitivity matrix};
    \item Optimizes a linear combination of spike
      perturbations to improve weak classes under constraints that
      protect strong ones and enforce a linear regime;
    \item Iterates with adaptive amplitude decay to converge
      toward a balanced accuracy profile.
\end{enumerate}

The method operates entirely post-hoc: it requires the trained model,
a small mini-batch of images for Hessian--vector products ($n = 256$),
and an evaluation set for sensitivity measurement. No full training set
access, no gradient-based optimization, and no backpropagation updates
are needed beyond the Hessian--vector products computations for eigendecomposition.

We validate the method on ResNet-50/CIFAR-10, demonstrating +7.5\,pp
accuracy on the weakest class (cat) and a 36\% reduction in class
accuracy standard deviation, with negligible impact on global accuracy
($-$0.2\,pp). Hessian Surgery outperforms focal loss fine-tuning
($\Delta\sigma = -0.76$\,pp) and class-balanced fine-tuning
($\Delta\sigma = -0.60$\,pp) in rebalancing, achieving
$\Delta\sigma = -3.05$\,pp without any retraining. We further provide
empirical evidence that class-sensitive directions are strongly aligned
with the spike subspace through directed bulk-walk experiments. On CIFAR-100, sequential deflated Surgery with 45 out of 99 spikes reduces $\sigma$ by 1.6\% on the held-out set.
On ISIC-2019, Hessian Surgery gains combine with Class-Balance improvements, reducing $\sigma$ by 12.3\% while improving balanced accuracy by 14.8\%.

\section{Background and Related Work}
\label{sec:background}

\subsection{Hessian Spectral Structure}
\label{sec:spectral-structure}

For a network with parameters $\vect{\theta} \in \R^p$ trained to
minimize $\mathcal{L}(\vect{\theta})$, the Hessian
$\Hess = \nabla^2 \mathcal{L}(\vect{\theta}) \in \R^{p \times p}$
is prohibitively large to compute or store
($p \approx 23 \times 10^6$ for ResNet-50).

Sagun et al.~\cite{sagun2017} first identified the bulk+outlier
structure of the Hessian spectrum in deep networks.
Ghorbani et al.~\cite{ghorbani2019} developed Stochastic Lanczos
Quadrature (SLQ) to estimate the full spectral density without
eigendecomposition, confirming the bulk+spike structure at scale.
Papyan~\cite{papyan2020} proved that the $C{-}1$ outlier eigenvalues
correspond to inter-class directions in the gradient feature space,
establishing a direct link between spectral structure and class
geometry. This structural result has been confirmed across
architectures~\cite{papyan2020,sagun2017}.

\begin{figure}[htbp]
    \centering    
    \includegraphics[width = \textwidth]{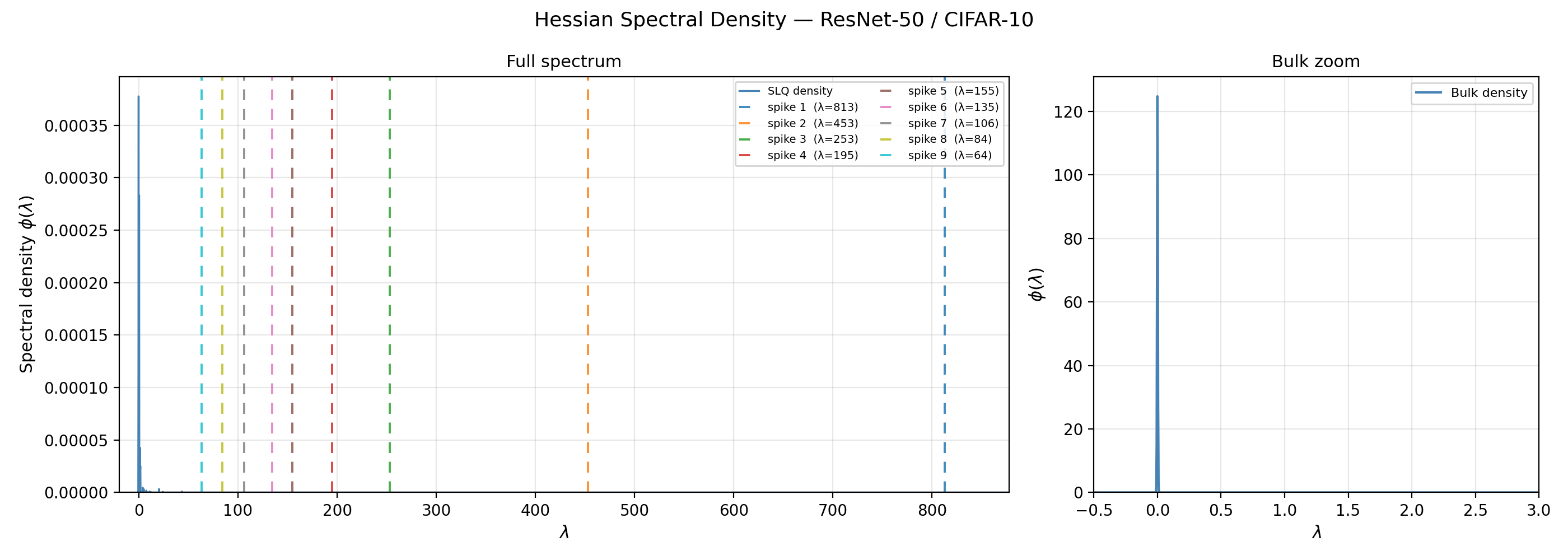}
    \caption{Hessian spectrum for the ResNet-50 architecture trained on CIFAR10. Full spectrum with dashed lines on each spike (left), zoom on the bulk (right).}
    \label{fig:schéma}
\end{figure}

\subsection{Approaches to Class Imbalance}

Existing methods for addressing per-class performance disparities
operate through the loss function or the data:
\begin{itemize}
    \item \textbf{Loss reweighting}: focal loss~\cite{lin2017},
      class-balanced loss~\cite{cui2019}, LDAM~\cite{cao2019};
    \item \textbf{Data-level}: oversampling, SMOTE, mixup variants;
    \item \textbf{Post-hoc calibration}: temperature scaling,
      Platt scaling-these adjust confidence, not accuracy.
\end{itemize}

To our knowledge, none of these methods operate on the Hessian spectrum
of the trained model.

Sharpness-aware methods like SAM~\cite{foret2021} and
GSAM~\cite{zhuang2022} perturb weights to seek flat minima, but they
operate in the full parameter space (worst-case loss direction) with no
class targeting. Our method is complementary: we perturb only along
spike eigenvectors with class-specific objectives.

\section{Experimental Setup}
\label{sec:setup}

\subsection{Model and Dataset}

We use a \textbf{ResNet-50}~\cite{he2016} trained on \textbf{CIFAR-10}
(50,000 training images, 10,000 test images, $32 \times 32$ RGB).
We initialize from ImageNet-pretrained weights and fine-tune the full
network on CIFAR-10, purely to avoid the cost of training ResNet-50
from scratch. Sagun et al.~\cite{sagun2017} and
Papyan~\cite{papyan2020} document the bulk+spike structure for
networks trained from random initialization on the target task and do
not specifically test the pretrained-then-fine-tuned regime; we extend
the empirical claim to that regime and verify it directly on our
trained model (\Cref{tab:eigenvalues}). The question of whether
pretraining biases the spike directions toward features inherited from
ImageNet, beyond what is needed for CIFAR-10 class structure, is open;
our experiments are consistent with the Papyan prediction
($C{-}1$ outliers correlated with class-mean directions) but do not
exclude residual pretraining-induced anisotropy. The model has 23,555,082
trainable parameters and achieves a baseline test accuracy of 84.9\%
with significant inter-class variance ($\sigma = 8.57\%$). Class-level
accuracies range from 68.6\% (cat) to 93.1\% (ship), reflecting the
difficulty disparity across classes.

\begin{table}[ht]
\centering
\caption{Baseline per-class accuracy of the trained ResNet-50 on
CIFAR-10.}
\label{tab:baseline}
\begin{tabular}{lcccccccccc}
\toprule
Class & plane & car & bird & cat & deer & dog & frog & horse & ship & truck \\
\midrule
Acc.\,(\%) & 89.2 & 90.7 & 81.7 & 68.6 & 85.2 & 69.0 & 90.6 & 90.5 & 93.1 & 89.5 \\
\bottomrule
\end{tabular}
\end{table}

We use the mean sparse categorical cross-entropy with softmax outputs:
\begin{equation}
\mathcal{L}(\vect{\theta}) = -\frac{1}{N} \sum_{i=1}^{N}
  \log \hat{p}_{y_i}(\vect{x}_i; \vect{\theta})
\label{eq:loss}
\end{equation}

We split the 10,000 test images into a 5,000-image
\emph{sensitivity set} (used for computing the sensitivity matrix
$\mat{S}$ and monitoring per-class accuracy during optimization) and
a 5,000-image \emph{held-out evaluation set} (used only for final
reporting). All accuracy figures reported in the results are measured
on the held-out set unless otherwise noted.

\section{Numerical Methods for Spectral Analysis}
\label{sec:numerical}

Computing the full Hessian $\Hess \in \R^{p \times p}$ with
$p \approx 23 \times 10^6$ is infeasible. We rely on matrix-free
methods that access $\Hess$ only through Hessian--vector products
(HVPs). This section presents both the algorithms and their convergence
guarantees, with additional details in \Cref{app:convergence}.

\subsection{Hessian--Vector Product (Pearlmutter Trick)}
\label{sec:hvp}

Given a vector $\vect{v} \in \R^p$, the product $\Hess\vect{v}$ can
be computed exactly via nested automatic
differentiation~\cite{pearlmutter1994}:
\begin{equation}
\Hess \vect{v} = \nabla_{\vect{\theta}}
  \bigl(\nabla_{\vect{\theta}} \mathcal{L} \cdot \vect{v}\bigr)
\label{eq:hvp}
\end{equation}
This requires two backpropagation passes and $O(p)$ memory-the same
as a single gradient computation.

In practice, the Hessian is estimated using a mini-batch of $n$ samples
rather than the full training set. Each HVP costs $O(p \cdot n)$. With
$n = 128$ images, a single HVP takes ${\sim}5$\,s on Apple Silicon CPU
for ResNet-50 ($p \approx 23 \times 10^6$).

\subsection{Lanczos Algorithm}
\label{sec:lanczos}

The Lanczos algorithm~\cite{lanczos1950} builds an orthonormal basis
$\{\vect{q}_1, \ldots, \vect{q}_m\}$ for the Krylov subspace
$\mathcal{K}_m(\Hess, \vect{v}) =
\mathrm{span}\{\vect{v}, \Hess\vect{v}, \ldots,
\Hess^{m-1}\vect{v}\}$ via a three-term recurrence
(\Cref{alg:lanczos}).

\begin{algorithm}[ht]
\caption{Lanczos tridiagonalization with full reorthogonalization}
\label{alg:lanczos}
\begin{algorithmic}[1]
\REQUIRE HVP oracle $\Hess\vect{v}$, starting vector $\vect{q}_1$
  ($\norm{\vect{q}_1}=1$), order $m$
\STATE $\beta_0 \leftarrow 0$, $\vect{q}_0 \leftarrow \vect{0}$
\FOR{$j = 1, \ldots, m$}
  \STATE $\vect{z} \leftarrow \Hess \vect{q}_j
    - \beta_{j-1}\vect{q}_{j-1}$
  \STATE $\alpha_j \leftarrow \vect{q}_j^\top \vect{z}$
  \STATE $\vect{z} \leftarrow \vect{z} - \alpha_j \vect{q}_j$
  \STATE \textbf{Reorthogonalize:}
    $\vect{z} \leftarrow \vect{z}
    - \sum_{i=1}^{j} (\vect{q}_i^\top \vect{z})\,\vect{q}_i$
    \hfill\COMMENT{Gram--Schmidt}
  \STATE $\beta_j \leftarrow \norm{\vect{z}}$
  \IF{$\beta_j < \varepsilon_{\mathrm{tol}}$}
    \STATE \textbf{break}
    \hfill\COMMENT{Krylov subspace exhausted}
  \ENDIF
  \STATE $\vect{q}_{j+1} \leftarrow \vect{z} / \beta_j$
\ENDFOR
\ENSURE Tridiagonal matrix
  $\mat{T}_m = \mathrm{tridiag}(\beta_{j-1}, \alpha_j, \beta_j)$,
  basis $\mat{Q}_m$
\end{algorithmic}
\end{algorithm}

The eigenvalues of $\mat{T}_m$ (\emph{Ritz values}) approximate the
extreme eigenvalues of $\Hess$, and the corresponding \emph{Ritz
vectors} $\tilde{\vect{v}}_i = \mat{Q}_m \vect{u}_i$ approximate the
true eigenvectors. The algorithm requires $m$ HVPs and stores $m$
vectors of dimension $p$, giving $O(mp)$ memory and
$O(m \cdot p \cdot n)$ computation. For $m=10$,
$p = 23 \times 10^6$, $n=128$: ${\sim}50$\,s total.

A critical point: the Lanczos algorithm does \emph{not} recover one
eigenvalue per iteration. Rather, all $m$ Ritz values converge
simultaneously to the extremal eigenvalues of $\Hess$, with convergence
rate governed by the spectral gaps. With $m = 10$, the 10 Ritz values
provide accurate approximations of the 9 spikes and the first bulk
eigenvalue-but this is because the gaps are large, not because
$m = 10 $. 

\subsection{Convergence Study}
\label{sec:convergence}

The Lanczos algorithm converges exponentially fast to well-separated
extremal eigenvalues, with a rate controlled by the spectral
gaps~\cite{saad1980}. For our spectrum (anisotropy ratio
$\lambda_{\max}/\lambda_{\mathrm{bulk}} \approx 62{,}000$), $m = 10$
iterations suffice to recover all 9 spikes reliably. The stochastic
Hessian approximation (mini-batch of $n$ samples) introduces
eigenvalue variance but preserves the subspace structure at $n \geq
128$: the mean principal angle between the top-9 spike subspace and
the reference ($n{=}512$) is $24.2\degree$ at $n{=}128$ and
$14.3\degree$ at $n{=}256$. We use $n{=}128$ on CIFAR-10 and CIFAR-100
(wall-clock: ${\sim}126$\,s per Lanczos run); on the more imbalanced
ISIC-2019 we increase to $n{=}256$ to reduce per-class HVP variance
(see \Cref{sec:isic-adaptations}). Full eigenvalue, eigenvector, subspace,
and timing tables are given in \Cref{app:convergence}.

\subsection{Stochastic Lanczos Quadrature (SLQ)}
\label{sec:slq}

To estimate the full spectral density
$\phi(t) = \frac{1}{p} \sum_{i=1}^{p} \delta(t - \lambda_i)$, we use
SLQ~\cite{ghorbani2019}, which combines the Hutchinson trace estimator
with Gauss--Ritz quadrature. For $\vect{v} \sim \mathcal{N}(0,
\frac{1}{p}\mat{I})$:
\begin{equation}
\phi_\sigma(t) \approx \E_{\vect{v}}\bigl[
  \vect{v}^\top f_\sigma(\Hess; t) \vect{v}
\bigr]
\approx \sum_{i=1}^{m} \omega_i \, f_\sigma(\ell_i; t)
\end{equation}
where $f_\sigma(\cdot; t)$ is a Gaussian kernel with bandwidth
$\sigma$, $\ell_i$ are the Ritz values, and
$\omega_i = U_{1,i}^2$ with $\mat{T}_m = \mat{U}\mat{\Lambda}\mat{U}^\top$.
We average over $k$ probe vectors for robustness (convergence
$\propto 1/\sqrt{k}$).

SLQ is used only for spectral density visualization in Figure \ref{fig:schéma}, not for Hessian
Surgery itself. We use $m=90$, $k=10$, $\sigma^2 = 10^{-5}$
(${\sim}30$\,min CPU) for full density estimation. 

\section{Spectral Validation}
\label{sec:validation}

\subsection{Empirical Verification of the Spike--Class Correspondence}

Papyan~\cite{papyan2020} predicts $C$ outlier eigenvalues stemming from
the $C$ class-mean directions in the gradient outer-product matrix
$\mat{G}$: the $C{-}1$ ``contrast'' directions yield large eigenvalues
clearly separated from the bulk, while the $C$-th direction (aligned
with the global mean) yields a noticeably smaller outlier whose
magnitude can be of bulk-level order. \Cref{tab:eigenvalues} reports the
top 10 Lanczos eigenvalues at our trained minimum and lets us check
this structure directly.

\begin{table}[ht]
\centering
\caption{Top 10 eigenvalues of the Hessian at the trained ResNet-50
minimum (single Lanczos run, $m{=}10$, $n{=}128$). 8 large eigenvalues
clearly separated from the bulk; a ninth borderline value $\sim 0$;
a tenth small negative value attributable to stochastic Lanczos noise
(see \Cref{app:convergence}). Absolute magnitudes fluctuate
across runs by a factor of ${\sim}1.5$ on the dominant spike;
ratios $\lambda_i/\lambda_1$ are stable.}
\label{tab:eigenvalues}
\begin{tabular}{ccccccccccc}
\toprule
Rank & 1 & 2 & 3 & 4 & 5 & 6 & 7 & 8 & 9 & 10 \\
\midrule
$\lambda$ & 828.6 & 577.8 & 310.7 & 243.5 & 153.2 & 112.5 & 58.9
  & 20.5 & $\sim$0 & $-$1.1 \\
\bottomrule
\end{tabular}
\end{table}

We read this table as follows. The 8 leading eigenvalues
($828.6 \to 20.5$) are unambiguous outliers: each is at least three
orders of magnitude above the median bulk eigenvalue
$\lambda_{\mathrm{bulk}} = 0.0134$, with anisotropy ratio
$\lambda_1/\lambda_{\mathrm{bulk}} \approx 6.2 \times 10^4$. The 9th
eigenvalue $\sim 0$ matches Papyan's prediction of a smaller $C$-th
outlier (global-mean direction): it is consistent with $C{-}1=9$
class-mean spikes overall, but its magnitude is in the bulk regime, so
operationally we treat it as a borderline direction rather than a
genuine spike. The 10th eigenvalue ($-1.1$) is due to stochastic Lanczos with
finite-batch HVPs ($n{=}128$) which produces noise of either sign at small
magnitudes whenever the true value is bulk-scale (\Cref{app:convergence}
documents this on repeated runs).

For Hessian Surgery we use $K = 9$ Lanczos directions: the 8 clear
spikes plus the 9th borderline direction. If the latter carries no
class-discriminative signal, the SLSQP solver assigns
$\alpha_9 \approx 0$; if it does, it is exploited. We never use the
$\sim$negative 10th direction.

\subsection{Bulk Walk: Class Invariance of the Bulk Subspace}
\label{sec:bulk-walk}

The spike--bulk decomposition implies that the complementary bulk
subspace should be inert with respect to class-level performance.
We verify this via a \emph{directed bulk walk}: 20 steps of size
$\varepsilon = 0.1$ in the Gram--Schmidt-projected bulk complement,
for a cumulative displacement $\|\delta\vect{\theta}\|/\|\vect{\theta}\| = 2.0$.
After 20 steps, the loss changes by $< 0.001$ and per-class accuracies
remain within $\pm 0.2\%$, while the direction stays orthogonal to
the spike subspace throughout (inner product $> 0.9999$). This experiment suggests class-discriminative
information is concentrated in 9 directions out of 23,555,082
parameters (${\sim}4 \times 10^{-7}$), which is precisely what makes
Hessian Surgery possible. Protocol, equations and full table are in
\Cref{app:bulk-walk}.

\section{Hessian Surgery: Method}
\label{sec:method}

\subsection{Spike--Class Sensitivity Matrix}
\label{sec:sensitivity}

\begin{definition}[Sensitivity matrix]
\label{def:sensitivity}
For spike eigenvectors $\{\vect{q}_1, \ldots, \vect{q}_K\}$ and probe
amplitude $\varepsilon > 0$, the sensitivity matrix
$\mat{S} \in \R^{K \times C}$ is:
\begin{equation}
S_{i,j} = \frac{
  \mathrm{acc}_j(\vect{\theta} + \varepsilon\,\vect{q}_i)
  - \mathrm{acc}_j(\vect{\theta} - \varepsilon\,\vect{q}_i)
}{2\varepsilon}
\label{eq:sensitivity}
\end{equation}
where $\mathrm{acc}_j(\vect{\theta})$ is the accuracy of class $j$ at
parameters $\vect{\theta}$.
\end{definition}

$S_{i,j} > 0$ means that moving along $+\vect{q}_i$ improves class
$j$. The matrix captures the finite-difference sensitivity of each
class's accuracy along each spike direction, providing a first-order
model for the effect of spike perturbations.

\begin{figure}[H]
\centering
\includegraphics[width=\textwidth]{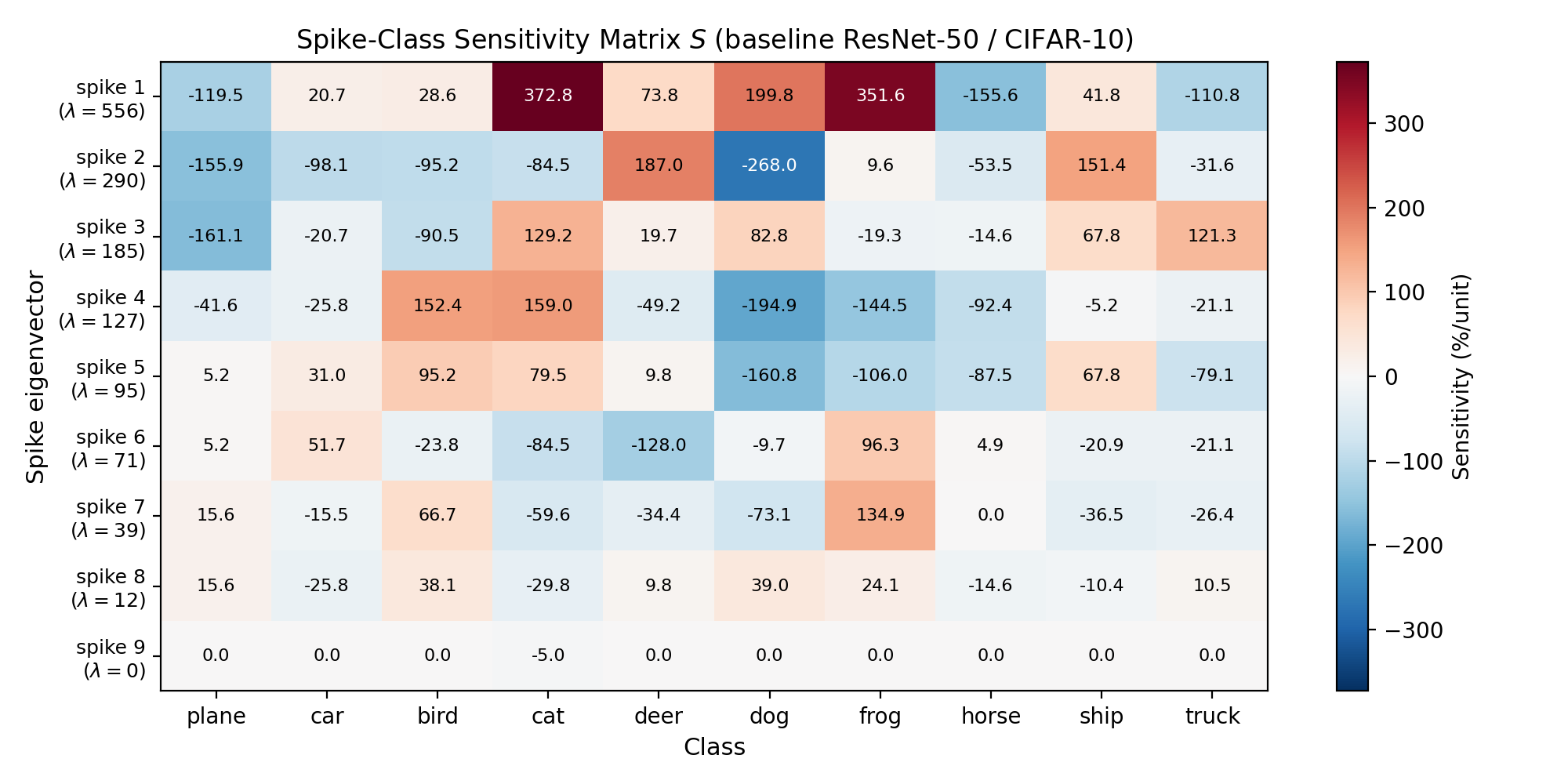}
\caption{Spike--class sensitivity matrix $\mat{S}$ for the baseline
ResNet-50/CIFAR-10 model ($\varepsilon = 0.02$, $n = 128$). Values
are in \%/unit of perturbation amplitude. Red: moving along
$+\vect{q}_i$ improves class $j$; blue: degrades it.}
\label{fig:sensitivity-matrix}
\end{figure}

Computing $\mat{S}$ requires $2K$ forward passes over the evaluation
set (one $+\varepsilon$ and one $-\varepsilon$ for each spike). The
probe amplitude $\varepsilon$ is logically separate from the optimization
constraint $\alpha_{\max}$ (Definition~\ref{def:sensitivity} is stated
for arbitrary $\varepsilon > 0$). At iteration~0 we set
$\varepsilon = 0.02$ to characterize the static sensitivity matrix
(\Cref{fig:sensitivity-matrix}); inside the iterative loop
(Algorithm~\ref{alg:spectral-surgery}) we tie $\varepsilon^{(t)} =
\alpha_{\max}^{(t)}$ so that the linearization is probed at the same
amplitude scale as the constraint, which keeps the sensitivity matrix
calibrated to the regime in which the SLSQP solver operates.

\Cref{fig:sensitivity-matrix} shows the sensitivity matrix computed at
the baseline model with $\varepsilon = 0.02$. The dominant spike
($\lambda_1 \approx 556$ on this run; absolute magnitudes vary
${\sim}\pm 50\%$ run-to-run, see \Cref{app:convergence}) is strongly
polarized: cat (+3.7), frog (+3.5),
and dog (+2.0) have large positive sensitivities, while plane ($-$1.2)
and horse ($-$1.6) have negative ones. This means that moving along
$+\vect{q}_1$ improves weak classes at the expense of some strong
classes-exactly the lever the optimization exploits.

The cat--dog anti-correlation is visible across multiple spikes:
spike~4 shows cat (+1.6) vs.\ dog ($-$1.9), while spike~2 shows
dog ($-$2.7) vs.\ deer (+1.9). This shared-spike coupling explains the
oscillatory convergence observed in \Cref{tab:iterations}.

\subsection{Constrained Coefficient Optimization}
\label{sec:optimization}

Given $\mat{S}$ and the current per-class accuracy vector
$\vect{a} \in \R^C$, we seek coefficients $\vect{\alpha} \in \R^K$
for the compound perturbation
$\delta\vect{\theta} = \sum_{i=1}^K \alpha_i \vect{q}_i$:
\begin{equation}
\begin{aligned}
\max_{\vect{\alpha}} \quad & \sum_{j=1}^C w_j \cdot
  \bigl(\mat{S}^\top \vect{\alpha}\bigr)_j \\
\text{s.t.} \quad
  & \norm{\vect{\alpha}} \leq \alpha_{\max} \\
  & (\mat{S}^\top \vect{\alpha})_j \geq -0.01 \quad
    \forall\, j : a_j > \alpha_{threshold}
\end{aligned}
\label{eq:optim}
\end{equation}

We optimize accuracy directly rather than loss because the algorithm
targets class-level balance. In the particular case of CIFAR-10, we take $\alpha_{threshold} = 0.85$, which characterizes a strong class that needs protecting across iterations.

\paragraph{Weights.}
The optimization weights $w_j$ are derived from the per-class error
$e_j = 1 - a_j$, raised to a power $p \geq 0$ and normalized:
\begin{equation}
w_j^{(p)} = \frac{e_j^p}{\sum_{k=1}^C e_k^p}
\label{eq:weights}
\end{equation}
The exponent $p$ controls the concentration of optimization pressure
on the weakest classes:
\begin{itemize}
  \item $p = 0$ (\emph{homogeneous}): uniform weights-all classes
    receive equal optimization pressure regardless of current accuracy. Tends to increase global accuracy and not emphasize standard deviation. 
  \item $p = 1$ (\emph{linear}, default): weight proportional to error,
    recovering the intuition that weaker classes deserve more correction.
  \item $p = 2$ (\emph{square}): weights proportional to squared error,
    concentrating pressure aggressively on the worst classes.
  \item $p = \nicefrac{1}{2}$ (\emph{sqrt}): softer concentration,
    reducing the gap between the weakest and second-weakest class.
\end{itemize}
The choice of $p$ interacts with both the error ratio across classes
and the optimization objective:

\begin{itemize}
  \item If $e_{\max}/e_{\min} < 5$ (mild-to-moderate regime, e.g.\
    CIFAR-10 with $e_{\mathrm{cat}}/e_{\mathrm{car}} \approx 4.4$):
    $p = 2$ (square) concentrates pressure on the dominant weak
    direction. This is most effective when the worst class is also
    coupled to a second weak class via a shared spike (e.g.\ cat$\leftrightarrow$dog
    on CIFAR-10), in which case focused gradient suppresses the
    oscillation that diffuse weighting would induce.
  \item If $e_{\max}/e_{\min} \geq 5$ (severe imbalance, e.g.\
    ISIC-2019 with $e_{\mathrm{VASC}}/e_{\mathrm{NV}} \approx 6.4$):
    $p = \nicefrac{1}{2}$ (sqrt) diffuses pressure across the long tail
    of weak classes; the $p \geq 1$ gradient is dominated by a single
    class whose sensitivity may be poorly estimated.
\end{itemize}

Note that the ``best'' $p$ also depends on the optimization target.
On the mild-to-moderate end of the spectrum, $p = 2$ minimizes
$\sigma$ when a shared spike couples the two weakest classes (the
CIFAR-10 case); otherwise $p = 1$ is a safer default. On the severe
end, $p = 0$ (homogeneous) maximizes the worst-class accuracy, while
$p = \nicefrac{1}{2}$ minimizes $\sigma$ by spreading pressure across
the tail.

\begin{table}[ht]
\centering
\caption{Recommended exponent $p$ (mode) for the class-error weighting
  as a function of imbalance severity and optimization target.}
\label{tab:p-rule}
\begin{tabular}{llll}
\toprule
Imbalance ($e_{\max}/e_{\min}$) & Minimize $\sigma$ & Maximize worst class & Note \\
\midrule
$< 5$ (mild-to-moderate) & $p=2$ (square) & $p=1$ (linear) & focus on coupled weak class \\
$\geq 5$ (severe) & $p=\nicefrac{1}{2}$ (sqrt) & $p=0$ (uniform) & avoid single-class domination \\
\bottomrule
\end{tabular}
\end{table}

\paragraph{Constraints.}
The norm constraint $\norm{\vect{\alpha}} \leq \alpha_{\max}$ ensures
we remain in the linear regime where $\mat{S}$ is a valid
approximation (see Appendix \ref{app:linearization}). The no-degradation constraint protects strong classes
($a_j > 85\%$) from losing more than 1\% per iteration.

We solve~\eqref{eq:optim} via Sequential Least-Squares Programming
(SLSQP), which handles the nonlinear norm constraint and linear
inequality constraints efficiently.

\subsection{Iterative Loop with Adaptive Decay}
\label{sec:loop}

The full procedure iterates: at each step, we recompute the Hessian
eigenvectors (since the spectrum changes after perturbation-the model
now sits at a different point in parameter space with different local
curvature), rebuild $\mat{S}$, solve for $\vect{\alpha}$, and apply
the perturbation. This recomputation is essential because a single
large perturbation would exceed the linear regime; instead, we take
moderate steps and re-linearize at each iteration.

\paragraph{Adaptive amplitude control.}
At each iteration $t$ we compute a scalar improvement signal $g_t$
(positive on success, negative on rollback) and maintain Adam
first/second moments with $\beta_1 = 1 - 4/T$, $\beta_2 = 1 - 1/T$,
where $T$ is the total number of Hessian Surgery iterations
budgeted for the run (effective EMA windows $T/4$ and $T$,
principled scaling to the iteration budget).
The bias-corrected SNR drives:
\begin{equation}
\alpha_{\max}^{(t)} = \alpha_{\min}
  + \frac{1 + \tanh(5 \cdot \mathrm{SNR}_t)}{2}
  \cdot \bigl(\alpha_{\max}^{(0)} - \alpha_{\min}\bigr).
\label{eq:adam-alpha}
\end{equation}
Full derivation, signal definition, and comparison with multiplicative
$\gamma$-decay are in \Cref{app:adam-alpha}.

\paragraph{Rollback mechanism.}
If an iteration increases $\sigma_t$ by more than 0.5\% (absolute)
or causes a per-class drop exceeding 7\%, we restore the previous
weights. The signal $g_t = -\Delta_{\max}$ propagates the failure
into the Adam moments, automatically reducing $\alpha_{\max}$ at the
next iteration.

\subsection{Complexity Analysis}

Each iteration of \Cref{alg:spectral-surgery} costs:
\begin{itemize}
    \item \textbf{Lanczos}: $m$ HVPs at $O(pn)$ each $= O(mpn)$.
      With $m{=}10$, $p{=}23{\times}10^6$, $n{=}128$: ${\sim}50$\,s.
    \item \textbf{Sensitivity}: $2K$ forward passes over the
      evaluation set. With $K{=}9$: ${\sim}90$\,s.
    \item \textbf{SLSQP}: negligible ($K{=}9$ variables).
\end{itemize}
Total per iteration: ${\sim}140$\,s. For $T{=}10$ iterations:
${\sim}25$\,min on a single CPU.

The complete algorithm is given in \Cref{alg:spectral-surgery}.

\bigskip

\begin{algorithm}[ht]
\caption{Hessian Surgery}
\label{alg:spectral-surgery}
\begin{algorithmic}[1]
\REQUIRE Trained model $\vect{\theta}_0$, evaluation set $(X, Y)$,
  HVP mini-batch $(X_h, Y_h)$
\REQUIRE $\alpha_{\max}^{(0)}, \alpha_{\min}, \beta_1, \beta_2,
  \varepsilon, K, T$
\STATE $\vect{a}_0 \leftarrow
  \mathrm{per\_class\_accuracy}(\vect{\theta}_0)$;
  $\sigma_0 \leftarrow \mathrm{std}(\vect{a}_0)$
\STATE $m \leftarrow 0$,\; $v \leftarrow 0$
  \hfill\COMMENT{Adam moments for SNR}
\FOR{$t = 1, \ldots, T$}
  \STATE $\{\lambda_i, \vect{q}_i\}_{i=1}^K \leftarrow
    \mathrm{Lanczos}(\Hess, m_{\mathrm{lanczos}})$
    \hfill\COMMENT{Recompute eigenvectors}
  \STATE $\mat{S} \leftarrow
    \mathrm{sensitivity}(\vect{\theta}, \{\vect{q}_i\}, \varepsilon^{(t)})$
    \hfill\COMMENT{Eq.~\eqref{eq:sensitivity}, $\varepsilon^{(t)} \leftarrow \alpha_{\max}^{(t)}$}
  \STATE $\vect{\alpha}^* \leftarrow
    \mathrm{solve\_SLSQP}(\mat{S}, \vect{a}_{t-1}, \alpha_{\max})$
    \hfill\COMMENT{Eqs.~\eqref{eq:optim},\eqref{eq:per-spike-budget}}
  \STATE $\vect{\theta}_{\mathrm{cand}} \leftarrow
    \vect{\theta}_{t-1} + \sum_i \alpha_i^* \vect{q}_i$
  \STATE $\vect{a}_{\mathrm{cand}} \leftarrow
    \mathrm{per\_class\_accuracy}(\vect{\theta}_{\mathrm{cand}})$
  \STATE $\Delta_{\max} \leftarrow
    \max_j\bigl([\vect{a}_{t-1}]_j - [\vect{a}_{\mathrm{cand}}]_j\bigr)$
  \IF{$\mathrm{std}(\vect{a}_{\mathrm{cand}}) > \sigma_{t-1} + 0.005$
      \textbf{ or } $\Delta_{\max} > 0.07$}
    \STATE $\vect{\theta}_t \leftarrow \vect{\theta}_{t-1}$;\;
      $\vect{a}_t \leftarrow \vect{a}_{t-1}$
      \hfill\COMMENT{Rollback}
    \STATE $g_t \leftarrow -\Delta_{\max}$
  \ELSE
    \STATE $\vect{\theta}_t \leftarrow \vect{\theta}_{\mathrm{cand}}$;\;
      $\vect{a}_t \leftarrow \vect{a}_{\mathrm{cand}}$
    \STATE $g_t \leftarrow \sigma_{t-1} - \mathrm{std}(\vect{a}_t)$
      \hfill\COMMENT{positive if improved}
  \ENDIF
  \STATE $\sigma_t \leftarrow \mathrm{std}(\vect{a}_t)$
  \STATE $m \leftarrow \beta_1 m + (1-\beta_1)g_t$;\;
    $v \leftarrow \beta_2 v + (1-\beta_2)g_t^2$
    \hfill\COMMENT{Adam moments, Eq.~\eqref{eq:signal}}
  \STATE $\mathrm{SNR}_t \leftarrow
    \tfrac{m/(1-\beta_1^t)}{\sqrt{v/(1-\beta_2^t)} + \varepsilon}$
  \STATE $\alpha_{\max} \leftarrow \alpha_{\min} +
    \tfrac{1 + \tanh(5\,\mathrm{SNR}_t)}{2}
    \bigl(\alpha_{\max}^{(0)} - \alpha_{\min}\bigr)$
    \hfill\COMMENT{Eq.~\eqref{eq:adam-alpha}}
\ENDFOR
\ENSURE Rebalanced model $\vect{\theta}_T$
\end{algorithmic}
\end{algorithm}

\newpage

\section{Results}
\label{sec:results}

In this section we provide the main results of the application of Hessian Surgery to CIFAR-10.

\subsection{Main Results: Class Rebalancing}

Table~\ref{tab:main-results} reports the per-class accuracy before and
after Hessian Surgery on the held-out set, using the
$\gamma$-decay run (10 scheduled iterations: 8 accepted iterations
plus 2 rollbacks at iterations 9--10; convergence trace in
\Cref{tab:iterations}). The Adam-controlled run (15 iterations,
\Cref{tab:comparison}) uses the same $\alpha_{\max}^{(0)}$ and
$\varepsilon$ but a different amplitude schedule, so the two runs
end at slightly different per-class accuracies; we report both
throughout.

\begin{table}[ht]
\centering
\caption{Per-class accuracy before and after Hessian Surgery
($\gamma$-decay run, 8 accepted iterations of 10 scheduled,
$\alpha_{\max}^{(0)} = 0.02$), measured on the held-out
evaluation set (5,000 images). The two weakest classes (cat, dog)
gain +7.5 and +5.9\,pp respectively.}
\label{tab:main-results}

{
\fontsize{12pt}{12pt}\selectfont     
\renewcommand{\arraystretch}{1} 
\setlength{\tabcolsep}{5pt}        

\begin{tabular}{lccc}
\toprule
Class & Baseline (\%) & After Surgery (\%) & $\Delta$ (pp) \\
\midrule
cat   & 68.6 & 76.1 & +7.5 \\
dog   & 69.0 & 74.9 & +5.9 \\
bird  & 81.7 & 83.2 & +1.5 \\
car   & 90.7 & 92.4 & +1.7 \\
truck & 89.5 & 88.4 & $-$1.1 \\
frog  & 90.6 & 89.2 & $-$1.4 \\
plane & 89.2 & 85.9 & $-$3.3 \\
ship  & 93.1 & 90.0 & $-$3.1 \\
horse & 90.5 & 84.0 & $-$6.5 \\
deer  & 85.2 & 81.7 & $-$3.5 \\
\midrule
Global & 84.9 & 84.7 & $\bm{-}$0.2 \\
\textbf{Std} & \textbf{8.57} & \textbf{5.52} & $\bm{-}$\textbf{3.05} \\
\bottomrule
\end{tabular}

}

\end{table}

\subsection{Convergence Dynamics}

\Cref{tab:iterations} shows the evolution across iterations. The
inter-class $\sigma$ decreases monotonically (with decay at
iteration~6 and rollbacks at iterations~9 and~10), while
$\alpha_{\max}$ adapts from 0.020 to 0.0069.

\begin{table}[ht]
\centering
\caption{Evolution per iteration (measured on the 5,000-image sensitivity
set). Decay at iteration~6; rollbacks at iterations 9 and 10.
The EMA smooths the decay signal.}
\label{tab:iterations}
\begin{tabular}{ccccccc}
\toprule
Iter & Global & $\sigma$ & EMA & $\alpha_{\max}$ & cat & dog \\
\midrule
0 (base) & 84.1 & 9.17 & 9.17 & 0.020 & 63.8 & 70.8 \\
1 & 84.9 & 7.05 & 8.53 & 0.020 & 70.4 & 75.4 \\
2 & 84.6 & 5.97 & 7.77 & 0.020 & 73.2 & 79.5 \\
3 & 85.0 & 6.22 & 7.30 & 0.020 & 75.1 & 73.3 \\
4 & 84.4 & 5.80 & 6.85 & 0.020 & 73.2 & 77.6 \\
5 & 85.1 & 6.12 & 6.63 & 0.020 & 75.5 & 73.5 \\
6 & 84.4 & 6.62 & 6.63 & 0.014 & 71.8 & 76.6 \\
7 & 84.8 & 5.52 & 6.30 & 0.014 & 76.3 & 75.2 \\
8 & 84.7 & 5.24 & 5.98 & 0.014 & 74.6 & 77.6 \\
9 & \multicolumn{6}{c}{\textit{rollback $\to$ decay:
  $0.014 \to 0.0098$}} \\
10 & \multicolumn{6}{c}{\textit{rollback $\to$ decay:
  $0.0098 \to 0.0069$}} \\
\bottomrule
\end{tabular}
\end{table}

Iterations 1--2 achieve most of the improvement (+9.4\% cat, +8.7\%
dog cumulative), making the process competitive with fine-tuning in
speed while offering direct control over each class's accuracy. Cat
and dog accuracies oscillate anti-symmetrically, driven by shared spike
eigenvectors with opposite sensitivities -consistent with their visual
similarity. The adaptive amplitude control dampens this oscillation
by reducing the perturbation budget after conflicting iterations.

\subsection{Bulk-walk experiment}
\label{app:bulk-walk}

The spike--bulk decomposition suggests a natural hypothesis: if the
$C{-}1$ spike eigenvectors encode inter-class structure, then the
complementary \emph{bulk subspace} (the remaining $p - (C{-}1)$
directions) should be inert with respect to class-level performance.
To test this, we perform a \emph{directed bulk walk}.

\paragraph{Protocol.} At step $t$, the walk proceeds as follows:
\begin{enumerate}
    \item \textbf{Recompute spike basis.} Run Lanczos at the current
      parameters $\vect{\theta}_t$ to obtain the top $K$ eigenvectors
      $\{\vect{q}_1^{(t)}, \ldots, \vect{q}_K^{(t)}\}$. This is
      necessary because the spike subspace may rotate as we move
      through parameter space.

    \item \textbf{Construct or update the walk direction.} If $t = 1$
      (or after a ``wall'' detection, see below), sample a random
      Gaussian vector $\vect{z} \sim \mathcal{N}(0, \mat{I}_p)$ and
      project it onto the bulk:
      \begin{equation}
      \vect{d}_{\mathrm{bulk}}^{(t)} = \vect{z}
        - \underbrace{\sum_{i=1}^{K}
          (\vect{q}_i^{(t)\top} \vect{z})\,
          \vect{q}_i^{(t)}}_{\text{remove spike components}}
        - \underbrace{\sum_{j \in \mathrm{hist}}
          (\vect{d}_j^\top \vect{z})\,
          \vect{d}_j}_{\text{remove past directions}}
      \label{eq:bulk-proj}
      \end{equation}
      The second Gram--Schmidt term removes all previously used walk
      directions, preventing the walk from degenerating into a random
      walk.

      If $t > 1$ and no wall is detected, re-project the current
      direction $\vect{d}_{\mathrm{bulk}}^{(t-1)}$ onto the updated
      spike complement and renormalize.

    \item \textbf{Wall detection.} If the re-projected direction has
      small norm ($\norm{\vect{d}'} < \tau$, with $\tau = 0.5$), the
      current direction has been ``absorbed'' by the spike subspace.
      In this case, we archive the old direction and sample a new one.

    \item \textbf{Step.} Apply
      $\vect{\theta}_{t+1} = \vect{\theta}_t + \varepsilon \cdot
      {\vect{d}}_{\mathrm{bulk}}^{(t)}$
      and measure loss, accuracy (global and per-class), and
      $\lambda_{\max}$.
\end{enumerate}

We execute 20 steps with $\varepsilon = 0.1$ for a cumulative
displacement $\norm{\delta\vect{\theta}} = \frac{\norm{\vect{\theta_f} -
\vect{\theta_i}}}{\vect{{\norm{\theta_i}}}} = 2.0$.

\begin{table}[ht]
\centering
\caption{Directed bulk walk: 20 steps with $\varepsilon = 0.1$,
cumulative $\norm{\delta\vect{\theta}} = 2.0$. Per-class accuracies
remain within $\pm 0.2\%$ of their starting values throughout.}
\label{tab:bulk-walk}
\begin{tabular}{lcc}
\toprule
Metric & Start & End (step 20) \\
\midrule
Loss & 0.5295 & 0.5293 \\
Accuracy & 84.98\% & 84.96\% \\
$\lambda_{\max}$ & 645.4 & 636.4 \\
$\langle \vect{d}, \mathrm{bulk}\rangle$ & - & 0.9999 \\
\bottomrule
\end{tabular}
\end{table}

\paragraph{Results and interpretation.}
After 20 steps ($\norm{\delta\vect{\theta}} = 2.0$), the loss changes
by $< 0.001$ and per-class accuracies remain within noise
($\pm 0.2\%$). The direction remains orthogonal to the spike subspace
throughout (inner product with the spike basis $> 0.9999$), confirming
numerical stability.

This result is telling: the model has moved substantially in parameter
space-a cumulative displacement of 2.0 in norm-yet its behavior is
indistinguishable from the starting point. The bulk subspace is
therefore a \emph{flat valley}: a vast region of parameter space where
the loss landscape is nearly constant.

The practical implication is that class-discriminative information is
concentrated in only 9 directions out of 23,555,082 parameters
(${\sim}4 \times 10^{-7}$ of the full space). This extreme
concentration is what makes Hessian Surgery possible: by operating in
the spike subspace, we can make targeted class-level changes without
disturbing the remaining directions.

\subsection{Sensitivity Matrix Structure}

The sensitivity matrix $\mat{S}$ (\Cref{fig:sensitivity-matrix})
reveals the spike--class coupling. For the dominant spike
($\lambda_1 \approx 556$), the sensitivity is strongly polarized: cat
(+3.7), frog (+3.5), and dog (+2.0) benefit from $+\vect{q}_1$,
while plane ($-$1.2) and horse ($-$1.6) are hurt. The sign convention
is arbitrary (Lanczos returns $\pm\vect{q}_i$), but the
\emph{structure} is robust: the weakest classes have the largest
magnitude sensitivities.

The dominant spike has the largest sensitivity magnitudes on the
weakest classes (cat, dog, frog), while the strongest classes
(ship, frog) are less affected - a pattern consistent with Papyan's
result that spikes encode inter-class directions, with the dominant
spike aligned toward the most separated classes in gradient feature
space~\cite{papyan2020}. The optimization in~\eqref{eq:optim}
exploits this structure across all 9 spikes simultaneously.

\subsection{Comparison with Training-Time Methods}
\label{sec:comparison}

Hessian Surgery operates entirely post-hoc. A natural question is how
it compares to standard methods that modify the training objective, and
to other truly post-hoc approaches. We compare against:
(i)~focal loss~\cite{lin2017} and class-balanced loss~\cite{cui2019},
both applied as fine-tuning from the same baseline (15 epochs,
lr $= 10^{-4}$);
(ii)~$\tau$-normalization~\cite{kang2020}, which rescales each
classifier column $w_c \leftarrow w_c / \|w_c\|^\tau$ (sweeping
$\tau \in \{0.25,\ldots,2.0\}$, reporting best);
(iii)~logit adjustment~\cite{menon2021}, which shifts logits by
$-\tau\log\pi_c$ at inference, where $\pi_c$ is the training class
frequency.

\begin{table}[ht]
\centering
\caption{Comparison of rebalancing methods on CIFAR-10 (ResNet-50,
held-out 5000 images). All start from the same baseline.
$\sigma$: standard deviation of per-class accuracy (held-out).
$\Delta\sigma$: change from baseline. Cat/dog are the two weakest
classes (68.6\%/69.0\% at baseline). The Hessian Surgery row uses
the Adam-controlled run (15 iterations); \Cref{tab:main-results} and
\Cref{tab:iterations} report the parallel $\gamma$-decay run, which
reaches a slightly different end-state ($\sigma=5.52$).}
\label{tab:comparison}
\begin{tabular}{lcccccc}
\toprule
Method & Global (\%) & $\sigma$ (\%) & $\Delta\sigma$ (pp)
  & Cat (\%) & Dog (\%) & Post-hoc? \\
\midrule
Baseline              & 84.9 & 8.57 & -   & 68.6 & 69.0 & - \\
Focal Loss FT         & 84.8 & 7.81 & $-$0.76 & 70.2 & 72.3 & No \\
Class-Balanced FT     & 84.4 & 7.97 & $-$0.60 & -  & -  & No \\
$\tau$-norm ($\tau=2$)& 84.7 & 7.90 & $-$0.67 & 74.8 & 66.3 & Yes \\
Logit Adjustment      & 84.9 & 8.57 & $\phantom{+}$0.00 & 68.6 & 69.0 & Yes \\
\textbf{Hessian Surgery} & \textbf{85.1}
  & \textbf{5.78} & $\bm{-}$\textbf{2.79} & \textbf{76.1} & \textbf{73.9} & Yes \\
FL + Hessian Surgery  & 85.1 & 5.51 & $-$3.06 & 79.1 & 73.5 & Yes$^\dagger$ \\
\bottomrule
\end{tabular}
\smallskip\par\noindent{\footnotesize
$^\dagger$ Post-hoc relative to the FL checkpoint; requires FL fine-tuning as a preprocessing step.}
\end{table}

Focal loss achieves a moderate reduction
($\Delta\sigma = -0.76$\,pp), improving the two weakest classes
(cat: $+1.6$\,pp, dog: $+3.3$\,pp) but causing degradations elsewhere
(horse: $-5.1$\,pp, ship: $-4.6$\,pp). This limited and uneven effect
is expected: focal loss down-weights well-classified \emph{samples},
not well-classified \emph{classes}. A class with high average accuracy
can still contain many hard samples, so the focal reweighting is only
indirectly correlated with class-level imbalance, offering no
guarantee of monotone improvement per class.

Class-balanced loss provides moderate rebalancing
($\Delta\sigma = -0.60$\,pp), as it explicitly compensates for class
frequency. However, it requires access to the full training set with
known class labels.

$\tau$-normalization achieves a modest reduction
($\Delta\sigma = -0.67$\,pp) but introduces a problematic trade-off:
the strongest $\tau$ (= 2.0) that best reduces $\sigma$ simultaneously
\emph{degrades} dog from 69.0\% to 66.3\% ($-$2.7\,pp) while
boosting cat to 74.8\%. This is a direct consequence of the absence
of any no-degradation constraint in the method: normalizing column norms
is a blunt instrument that cannot distinguish between classes that share
similar weight norms for unrelated reasons.

Logit adjustment is \emph{exactly equivalent to the baseline} on
CIFAR-10 ($\Delta\sigma = 0.00$\,pp). CIFAR-10 is perfectly balanced
($\pi_c = 0.1$ $\forall c$), so $\log\pi_c$ is identical for all
classes and the adjustment cancels in the softmax. The method has no
leverage when there is no prior imbalance to correct.

Hessian Surgery achieves the largest reduction
($\Delta\sigma = -2.79$\,pp, Adam, 15\,iter) while operating entirely
post-hoc, with no training data beyond 256 images for Hessian
estimation, and with a built-in no-degradation constraint that prevents
the dog-style sacrifice seen in $\tau$-normalization.

\subsection{Omega Mode Ablation}
\label{sec:omega-ablation}

We evaluate the three non-trivial weighting modes ($p \in \{\nicefrac{1}{2}, 1, 2\}$,
\Cref{tab:p-rule}) by running 15 iterations of Hessian Surgery on the
CIFAR-10 baseline and 10 iterations on the ISIC-2019 baseline. All runs
use the same $\alpha_{\max}$ and constraint parameters; only $p$ varies.
The ISIC-2019 dataset (8-class skin-lesion classification, severely
imbalanced with class frequencies ranging from 50.8\% for the majority
class NV to 0.9\% for VASC) is introduced in detail in
\Cref{sec:isic}; we include it here to test the $p$-rule beyond
the mildly imbalanced CIFAR-10 setting.

\begin{table}[ht]
\centering
\caption{Omega mode ablation on CIFAR-10 (held-out 5000 images, 15 iter)
and ISIC-2019 (test set 1900 images, 10 iter), both on the cross-entropy
baseline. Best per dataset in bold.}
\label{tab:omega-ablation}
\begin{tabular}{llcccc}
\toprule
Dataset & Mode ($p$) & Global (\%) & $\sigma$ (\%) & $\Delta\sigma$ (pp)
  & Weakest classes \\
\midrule
\multirow{4}{*}{CIFAR-10}
  & Baseline     & 84.9 & 8.57 & -   & cat 68.6, dog 69.0 \\
  & sqrt ($\nicefrac{1}{2}$) & 85.0 & 6.06 & $-$2.51 & cat 75.7, dog 72.5 \\
  & linear ($1$) & 85.2 & 5.96 & $-$2.61 & cat 73.4, dog 76.6 \\
  & \textbf{square ($2$)} & \textbf{84.9} & \textbf{5.69} & $\bm{-}$\textbf{2.88}
    & cat \textbf{76.9}, dog 73.9 \\
\midrule
\multirow{4}{*}{ISIC-2019}
  & Baseline     & 67.7 & 23.6 & -   & VASC 10.5, DF 33.3 \\
  & \textbf{sqrt ($\nicefrac{1}{2}$)} & 69.9 & \textbf{18.4} & $\bm{-}$\textbf{5.2}
    & VASC \textbf{31.6}, DF \textbf{61.1} \\
  & linear ($1$) & \textbf{71.4} & 21.5 & $-$2.1 & VASC 21.1, DF 50.0 \\
  & square ($2$) & 70.5 & 19.5 & $-$4.1 & VASC 26.3, DF 61.1 \\
\bottomrule
\end{tabular}
\end{table}

The two datasets exhibit \emph{opposite} optimal modes, which is exactly
the prediction of the selection rule in \Cref{tab:p-rule}.

\paragraph{CIFAR-10: square wins.}
The error ratio $e_{\max}/e_{\min} \approx 4.4$ (cat at $32.3\%$ error,
car at $7.3\%$) lies in the mild-to-moderate regime. Concentrating pressure on cat
(square, $p=2$) achieves the best $\sigma$ ($-$2.88\,pp) because the
shared spike coupling with dog - which drives oscillations - is suppressed
by the focused gradient. Under sqrt, the optimization spreads weight across
all weak classes simultaneously, which amplifies the cat$\leftrightarrow$dog
anti-correlation (3 rollbacks vs.\ 1 for square). Under linear,
convergence is smooth but less effective on the dominant direction.

\paragraph{ISIC-2019: sqrt wins.}
The error ratio $e_{\mathrm{VASC}}/e_{\mathrm{NV}} \approx 6.4$ places
ISIC in the severe regime. Under linear ($p=1$), VASC (error
$\approx 89\%$) absorbs disproportionate optimization pressure at the
expense of AK, BKL, and SCC, yielding the worst balanced accuracy
(49.8\%). Under square, VASC monopolization worsens further. The sqrt
mode compresses the error distribution sufficiently to spread pressure
across all underperforming classes, achieving the best $\sigma$
($-$5.2\,pp) and balanced accuracy (52.5\%), with both DF and VASC
gaining substantially ($+$27.8\,pp and $+$21.1\,pp from baseline).
Linear and square both exhibit stronger plateauing after iteration~5,
while sqrt converges monotonically for 10 iterations.

\subsection{Bulk Fine-Tuning Experiment}
\label{sec:bulk-ft}

We tested whether the spike-rebalanced CIFAR-10 model can be further
improved via fine-tuning projected into the bulk subspace
($\vect{g}_{\mathrm{bulk}} = \vect{g} -
\mat{Q}\mat{Q}^\top\vect{g}$). \textbf{Protocol.} Starting from the
post-Surgery checkpoint of \Cref{tab:main-results}, we fine-tune for
15 epochs on the full CIFAR-10 training set (50{,}000 images,
batch~64, Adam, lr~$=10^{-4}$, cross-entropy loss). At each gradient
step we project the gradient orthogonally to the current top-9 spike
subspace; the spikes $\mat{Q} \in \R^{p \times 9}$ are recomputed
once per epoch via Lanczos ($m{=}30$, $n_{\mathrm{HVP}}{=}128$) on the
current weights, since the spike subspace rotates as training proceeds.
A train/sensitivity 90/10 split is held out from the projection batches.
The classic-FT comparison uses identical hyperparameters with the
projection disabled.

\begin{table}[h]
\centering
\caption{Fine-tuning after Hessian Surgery (15 epochs, lr=$10^{-4}$).
Bulk FT: gradients projected $\perp$ spikes, recomputed every epoch.
Classic FT: standard fine-tuning. Both methods erode the Surgery gains.}
\label{tab:bulk-ft}
\begin{tabular}{lccc}
\toprule
 & Post-Surgery & Bulk FT & Classic FT \\
\midrule
Cat acc. & 76.1\% & 71.0\% ($-$5.1) & 69.2\% ($-$6.9) \\
Dog acc. & 74.9\% & 71.5\% ($-$3.4) & 68.2\% ($-$6.7) \\
Global & 84.7\% & 85.1\% (+0.4) & 84.9\% (+0.2) \\
$\sigma$ & 5.52\% & 7.53\% (+2.01) & 8.64\% (+3.12) \\
\bottomrule
\end{tabular}
\end{table}

Even with spike recomputation at every epoch, bulk-projected
fine-tuning erodes most of the Surgery gains ($\sigma$: $5.52 \to
7.53\%$, +2.01\,pp). Classic fine-tuning is worse still
($\sigma \to 8.64\%$), nearly reverting to baseline dispersion
($8.57\%$). The spike subspace \emph{rotates} during training,
rendering the projection stale within each epoch. Both methods improve
global accuracy slightly (+0.2--0.4\,pp), confirming that the bulk
contains useful signal for overall performance, but this comes at the
cost of the class-rebalancing achieved by Surgery.
This establishes that Hessian Surgery is best applied as a
\emph{terminal} post-hoc step, not as a precursor to further training.

\section{Scalability: CIFAR-100}
\label{sec:cifar100}

CIFAR-100 provides a natural scalability test with $C = 100$ classes
(grouped into 20 superclasses), 500 training images per class (vs.\
5,000 for CIFAR-10), and identical image resolution ($32 \times 32$).
Papyan's theory predicts $C{-}1 = 99$ spike eigenvalues, making the
full spike subspace significantly larger than in the CIFAR-10 setting.

\subsection{Setup and Memory Constraints}

We apply Hessian Surgery to the same ResNet-50 architecture fine-tuned
on CIFAR-100, achieving a baseline accuracy of 60.3\% with inter-class
standard deviation $\sigma = 16.91\%$ on the sensitivity set (16.28\%
on the held-out set; the small discrepancy reflects sample variance at
500 samples per class), substantially higher than CIFAR-10
($\sigma = 8.57\%$). All CIFAR-100 numbers in this section are reported
on the sensitivity set unless explicitly labelled \emph{held-out}; the
held-out set is used only for the final comparison
(\Cref{tab:cifar-comparison}).

The main challenge is memory: recovering all 99 spikes would require
Lanczos with $m \geq 100$, storing 100 vectors of dimension
$p = 23.5 \times 10^6$ in float32, totaling ${\sim}9.4$\,GB for the
Lanczos basis alone. On our hardware (16\,GB RAM), this exceeds the
available memory once the model, gradients, and HVP intermediates are
accounted for.

\subsection{Sequential Deflated Hessian Surgery}

To access spikes beyond the memory limit, we propose
\emph{sequential deflated Hessian Surgery}. Instead of a single
Lanczos run with $m = 100$, we run multiple phases that each target a
different slice of the spike spectrum using deflation.

Let $Q_{\mathrm{prev}} \in \R^{p \times k}$ denote the accumulated
eigenvectors from all previous phases. At phase $\ell \geq 2$, we
define a \emph{deflated HVP oracle}:
\begin{equation}
\widetilde{\Hess}\vect{v}
  = (\mat{I} - Q_{\mathrm{prev}} Q_{\mathrm{prev}}^\top)\,
    \Hess\,
    (\mat{I} - Q_{\mathrm{prev}} Q_{\mathrm{prev}}^\top)\,
    \vect{v}
\label{eq:deflated-hvp}
\end{equation}
which projects out the already-treated spike directions from both the
input and the output of the HVP. The double-sided projection ensures
the deflated operator remains symmetric, as required by Lanczos. In
this deflated spectrum, the previously dominant spikes have been
replaced by zeros, and the next group of spikes becomes the extremal
eigenvalues-potentially improving Lanczos convergence through better
gap ratios.

Each phase runs Lanczos with $m = 20$ on $\widetilde{\Hess}$ to
recover 15 new spikes, then applies Hessian Surgery (3 iterations)
using these directions. The accumulated eigenvectors
$Q_{\mathrm{prev}}$ are re-orthogonalized via QR before each new
phase. At phase $\ell$, memory usage is $O(mp)$ for the Lanczos basis
plus $O(15(\ell{-}1) \cdot p)$ for $Q_{\mathrm{prev}}$-substantially
less than a single Lanczos run with $m = 100$.

\paragraph{Adaptive amplitude scaling.}
A fixed $\alpha_{\max}$ is suboptimal across phases: the eigenvalue
magnitudes decrease from $\lambda \approx 600$ (phase~1) to
$\lambda \approx 20$ (phase~3), so a perturbation of fixed norm in
parameter space produces a much smaller effect on the loss landscape
in later phases. We scale $\alpha_{\max}$ inversely with
$\sqrt{\lambda_{\max}}$ to maintain a roughly constant perturbation
budget across phases:
\begin{equation}
\alpha_{\max}^{(\ell)} = \alpha_{\max}^{(1)} \cdot
  \sqrt{\frac{\lambda_{\max}^{(1)}}{\lambda_{\max}^{(\ell)}}}
\label{eq:adaptive-alpha}
\end{equation}

\subsection{Results}

\Cref{tab:cifar100-phases} summarizes the evolution across 3 phases
(measured on the sensitivity set). The best inter-class standard
deviation is reached at the end of phase~3 (spikes 1--45). Phase~4
(spikes 46--60) could not be completed due to memory constraints
(eigenvector storage exceeding 2\,GB).

\begin{table}[ht]
\centering
\caption{Sequential deflated Hessian Surgery on CIFAR-100 (sensitivity
set). Each phase recovers 15 new spikes via deflated Lanczos and runs
3 Surgery iterations. Best $\sigma$ at end of phase~3.}
\label{tab:cifar100-phases}
\begin{tabular}{lcccccc}
\toprule
Phase & Spikes & $\lambda_{\max}$ & $\alpha_{\max}$
  & Global (\%) & $\sigma$ (\%) & $\Delta\sigma$ (pp) \\
\midrule
Baseline & - & - & - & 60.3 & 16.91 & - \\
Phase 1 & 1--15 & 442.0 & 0.029 & 60.2 & 16.40 & $-$0.51 \\
Phase 2 & 16--30 & 83.9 & 0.067 & 60.9 & 16.49 & $-$0.42 \\
Phase 3 & 31--45 & 24.1 & 0.125 & \textbf{60.8} & \textbf{15.86}
  & $\bm{-}$\textbf{1.06} \\
\bottomrule
\end{tabular}
\end{table}

Deflation quality is verified by the cross-correlation between
eigenvectors of successive phases: $\max |Q_\ell^\top Q_{\ell'}| <
4 \times 10^{-5}$ throughout, confirming that each phase recovers
genuinely new directions.

\paragraph{Spectral hierarchy.}
Papyan~\cite{papyan2020} establishes that the $C{-}1$ spike eigenvectors
correspond to inter-class directions in the gradient outer-product
matrix, ordered by the magnitude of inter-class contrast they encode.
The phase results are consistent with this: phase~1 (dominant spikes,
$\lambda \approx 440$) produces a moderate $\sigma$ reduction with a
slight global accuracy drop, while phases~2--3 (intermediate spikes,
$\lambda \approx 24$--$84$) continue the rebalancing for a cumulative
$-$1.06\,pp from baseline. Whether the progressively weaker spikes
encode systematically narrower class contrasts-or simply carry less
curvature and therefore smaller per-step effects-cannot be
determined from these three phases alone.

Phase~4 could not be completed due to memory constraints, but the
diminishing eigenvalue magnitudes suggest decreasing returns.

\begin{figure}[H]
\centering
\includegraphics[width=18cm]{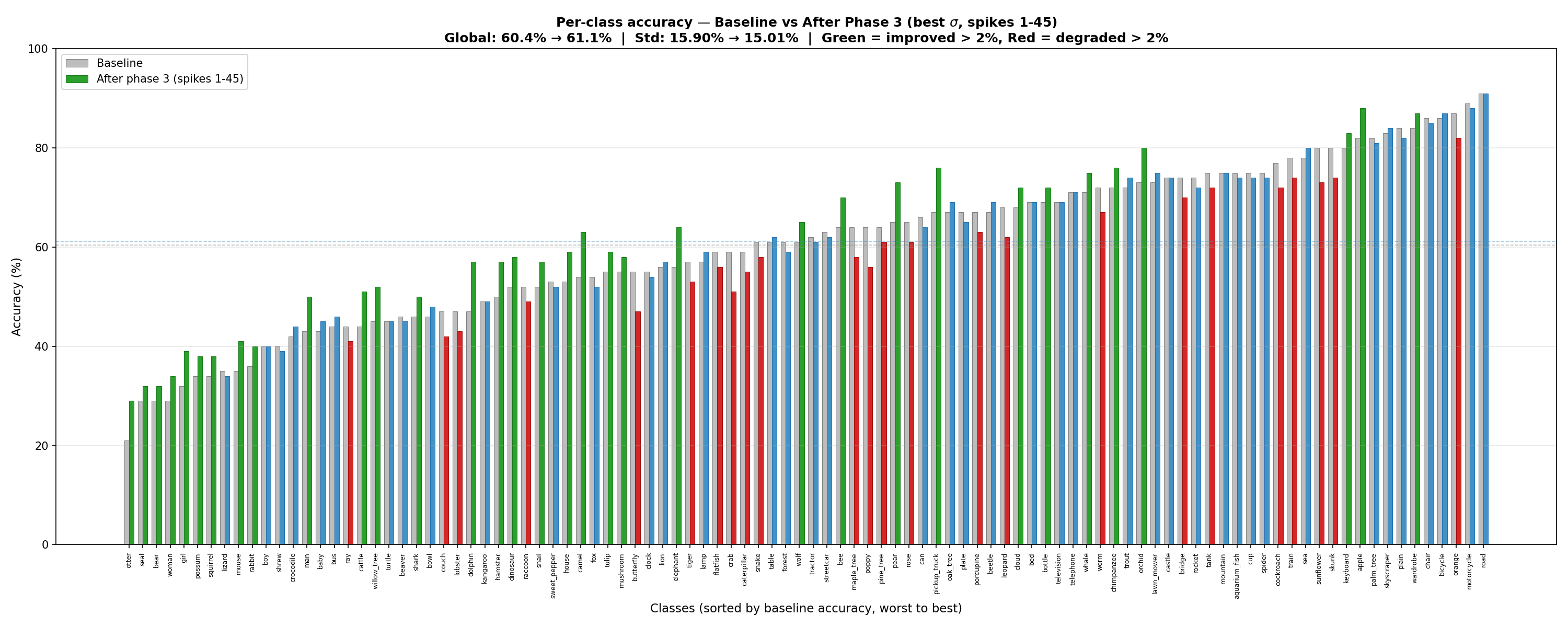}
\caption{Per-class accuracy on CIFAR-100 before (grey) and after
deflated Hessian Surgery through phase~3, using 45 spikes. Green:
improvement $> 2$\,pp; red: degradation $> 2$\,pp. Classes sorted by
baseline accuracy.}
\label{fig:cifar100-classes}
\end{figure}

\paragraph{Per-class analysis.}
\Cref{fig:cifar100-classes} shows the per-class accuracy before and
after deflated Surgery (at the phase~3 checkpoint). Gains are
concentrated on weak classes: willow tree (+14.6\,pp), woman (+12.0\,pp),
shrew (+9.1\,pp), while degradations are spread across strong classes
(plate $-$14.6\,pp, bridge $-$11.8\,pp). Three classes remain below
30\% (baby, bear, otter), possibly because their discriminative
structure resides in spikes beyond
the 45 we accessed.

\newpage

\paragraph{Decile analysis.}
To quantify the redistribution pattern beyond individual classes,
\Cref{tab:deciles} groups the 100 classes into deciles by baseline
accuracy and reports the mean accuracy change after Surgery.

\begin{table}[H]
\centering
\caption{Mean accuracy change by baseline decile on CIFAR-100 after
deflated Hessian Surgery (phase~3, 45 spikes), measured on the
held-out set.}
\label{tab:deciles}
\begin{tabular}{lcc}
\toprule
Decile & Baseline acc.\,(\%) & $\Delta$ Phase 3 (pp) \\
\midrule
1 (weakest)    & 30.0 & +2.2 \\
2              & 41.8 & $-$0.1 \\
3              & 50.8 & $-$1.3 \\
4              & 54.9 & +5.1 \\
5              & 59.7 & +0.3 \\
6              & 64.3 & $-$0.8 \\
7              & 67.8 & $-$0.9 \\
8              & 73.4 & $-$1.4 \\
9              & 77.7 & $-$2.0 \\
10 (strongest) & 86.2 & $-$0.6 \\
\bottomrule
\end{tabular}
\end{table}

The weakest decile gains +2.2\,pp and the upper deciles (8--10)
show moderate degradation ($-$0.6 to $-$2.0\,pp), consistent
with a net transfer from strong to weak classes. Decile~4 shows
an anomalous +5.1\,pp gain that exceeds the weakest decile-a
reminder that with only 45 out of 99 spikes and limited per-class
samples, the redistribution pattern at this scale is noisy and should
not be over-interpreted.

\paragraph{Comparison with CIFAR-10.}
\Cref{tab:cifar-comparison} compares the two datasets on their
respective held-out evaluation sets. The relative $\sigma$ reduction on
CIFAR-100 (1.6\%) is modest compared to CIFAR-10 (35.6\%), which is
expected given that we access only 45 out of 99 spikes (45\% of the
subspace, vs.\ 100\% on CIFAR-10). The gap between sensitivity-set
results ($-$6.2\% relative) and held-out results ($-$1.6\%) suggests
partial overfitting to the sensitivity set, which is more pronounced
with 100 classes and limited per-class samples.

\begin{table}[ht]
\centering
\caption{Hessian Surgery results across datasets.}
\label{tab:cifar-comparison}
\begin{tabular}{lccccc}
\toprule
Dataset & $C$ & Spikes used & $\sigma$: before $\to$ after
  & Relative $\Delta\sigma$ & $\Delta$ acc. \\
\midrule
CIFAR-10 & 10 & 9/9 (100\%)
  & 8.57\% $\to$ 5.52\% & $-$35.6\% & $-$0.2\% \\
CIFAR-100 & 100 & 45/99 (45\%)
  & 16.28\% $\to$ 16.02\% & $-$1.6\% & +0.1\% \\
ISIC-2019$^\dagger$ & 8 & 7/7 (100\%)
  & 23.6\% $\to$ 16.4\% & $-$30.5\% & $-$0.2\% \\
\bottomrule
\end{tabular}
\smallskip\par\noindent{\footnotesize $^\dagger$ FL+HS condition: Surgery applied after focal loss pre-training.
$\Delta$ acc.\ relative to the FL checkpoint ($-$0.2\,pp vs.\ FL global accuracy of 72.5\%).}
\end{table}

\newpage

\section{Real-World Imbalanced Medical Imaging: ISIC-2019}
\label{sec:isic}

CIFAR-10 and CIFAR-100 provide clean, balanced benchmarks for spectral
analysis. We now evaluate Hessian Surgery on a more demanding real-world
setting: the ISIC-2019 skin lesion dataset, characterized by severe class
imbalance and visually confusable categories.

\subsection{Dataset and Challenges}

ISIC-2019 contains 25,331 dermoscopic images across 8 diagnostic
categories: melanoma (MEL), melanocytic nevi (NV), basal cell carcinoma
(BCC), actinic keratosis (AK), benign keratosis-like lesions (BKL),
dermatofibroma (DF), vascular lesions (VASC), and squamous cell carcinoma
(SCC). Class frequencies span two orders of magnitude: NV comprises
$\sim$67\% of training images, while VASC and DF together account for
$<$1\%. A ResNet-50 trained with standard cross-entropy on this dataset
accordingly develops a strongly polarized accuracy profile: NV reaches
85.9\% while VASC drops to 10.5\% and DF to 33.3\% on the held-out test
set (Table~\ref{tab:isic-results}).

Unlike CIFAR-10 where performance gaps reflect genuine visual difficulty
(cats and dogs are hard to distinguish), ISIC-2019 mixes two qualitatively
different sources of imbalance. Some classes-notably VASC and DF-are
\emph{rare but visually distinctive}: a dermatofibroma has a characteristic
white central scar, vascular lesions display unmistakable reddish
vessel-like patterns. Their low accuracy is a data scarcity artifact, not
a representational failure. Other classes-MEL versus NV, or SCC versus
BCC-are \emph{frequent but visually confusable}: both pairs share
overlapping pigmentation patterns that place them close in the learned
feature space, creating genuine discriminative ambiguity. This distinction
has direct implications for what Hessian Surgery can achieve.

By Papyan's result~\cite{papyan2020}, the ISIC-2019 model should exhibit
$C-1 = 7$ Hessian spikes. We verify this and use all 7 spikes throughout.

\subsection{Adaptations to the Severely Imbalanced Setting}
\label{sec:isic-adaptations}

Two modifications of the baseline Hessian Surgery procedure are required
to handle ISIC-2019's imbalance. The Adam-style amplitude control
(\Cref{eq:adam-alpha}) is unchanged from the general method and was also
applied to the CIFAR-10 and CIFAR-100 experiments.

\paragraph{Stratified sensitivity set.}
In CIFAR-10, any random mini-batch provides reliable per-class gradients
because all classes are equally represented. With ISIC-2019, an
unrestricted sample would contain $\sim$1000 NV images and $\sim$2 VASC
images per 1500-image batch, making the sensitivity matrix $\mat{S}$
extremely noisy for rare classes. We construct a stratified sensitivity
set by sampling at most 250 images per class from the training split,
yielding 1918 images in total (DF: 203, VASC: 215 since all available
training images are included). This ensures that each class contributes
a comparable signal to the sensitivity estimation, without introducing
any evaluation-time data leakage since the sensitivity set is drawn from
the training split.

\paragraph{HVP batch size.}
For ISIC-2019, the Hessian--vector products entering the Lanczos
iteration use $n = 256$ samples (vs.\ $n = 128$ on CIFAR-10). The FL+HS run was done using $n=64$ samples. The
HVP composition is matched to the loss the checkpoint was trained
with, so that we estimate the Hessian of the actual training
objective at its minimum:
\begin{itemize}
  \item For the cross-entropy and focal-loss checkpoints (HS alone,
    FL+HS), the 256 samples are drawn \emph{uniformly} from the
    training split, inheriting the natural class frequencies. This
    estimates $\nabla^2 \mathcal{L}_{\mathrm{CE/FL}}$ integrated over
    the training distribution.
    \item For the class-balanced checkpoint (CB+HS), the 256 samples are
    drawn \emph{stratified, 32 per class}. We use it as a
    pragmatic surrogate that is much closer to the CB-minimum operator
    than the natural-distribution CE Hessian would be.
\end{itemize}
The larger sample size (256 vs.\ 128) compensates for the
heavier-tailed bulk spectrum on ISIC and yields more stable spike
eigenvectors across runs. Stratification of the sensitivity set
(previous paragraph) is logically separate: it concerns the per-class
accuracy probes used to build $\mat{S}$, not the operator whose
spectrum is being estimated.

\paragraph{Per-spike amplitude bounds.}
On ISIC-2019, eigenvalues span roughly one decade
($\lambda_1 \approx 600$, $\lambda_7 \approx 20$), making a single
global L2 budget inappropriate. We replace it with per-component box
constraints
\begin{equation}
  |\alpha_i| \leq \alpha_{\max} \sqrt{\frac{\lambda_{\min}}{\lambda_i}},
  \quad i = 1, \ldots, K,
  \label{eq:per-spike-budget}
\end{equation}
ensuring each perturbation induces a constant second-order cost
$\alpha_i^2 \lambda_i \lesssim \alpha_{\max}^2 \lambda_{\min}$.
Derivation and comparison with the CIFAR-100 inter-phase scaling are
in \Cref{app:per-spike}.

\paragraph{Class-error weighting exponent.}
On CIFAR-10, the weakest class (cat) sits at $a_j \approx 67.7\%$
and the strongest (car) at $a_j \approx 92.7\%$, giving an error
ratio $e_{\max}/e_{\min} \approx 4.4$. Per the rule of
\Cref{tab:p-rule}, this places CIFAR-10 in the mild-to-moderate
regime where $p = 2$ (square) minimizes $\sigma$ by concentrating
pressure on the dominant weak direction.

On ISIC-2019 the situation is qualitatively different. With NV at
$85.9\%$ and VASC at $10.5\%$ in the baseline, the error ratio
reaches $e_{\mathrm{VASC}}/e_{\mathrm{NV}} \approx 6.4$. Under $p = 1$,
VASC and DF absorb nearly all optimization pressure, which can cause
the surgery to overshoot on these two classes at the expense of
moderate-error classes (AK, BKL, SCC) that also need improvement.

The \emph{square} mode ($p = 2$) further amplifies this imbalance and
produces oscillations. The \emph{sqrt} mode ($p = \nicefrac{1}{2}$)
compresses the error distribution, spreading pressure more evenly across
all underperforming classes. We use $p = \nicefrac{1}{2}$ for the
ISIC-2019 experiments, which provides more stable convergence across
the full class spectrum without sacrificing the large gains on VASC
and DF.

\subsection{Composition with Training-Time Methods}
\label{sec:isic-composition}

Hessian Surgery is a post-hoc method applied to a fixed trained model.
On ISIC-2019, the extreme class imbalance makes the cross-entropy
baseline a poor starting point: a model trained with standard
cross-entropy devotes almost no representational capacity to VASC or
DF. We therefore consider three training-time starting points and the
corresponding compositions with Surgery:
\begin{itemize}
    \item \textbf{HS} (on CE baseline): Surgery applied directly to the
      cross-entropy baseline.
    \item \textbf{FL} and \textbf{FL+HS}: a focal-loss fine-tuning
      ($\gamma = 2$, 15 epochs,
      following~\cite{lin2017}) used either alone or as a starting
      point for Surgery.
    \item \textbf{CB} and \textbf{CB+HS}: a class-balanced
      fine-tuning~\cite{cui2019} (effective-number reweighting,
      $\beta = (N-1)/N$ for $N$ the training-set size,
      multi-phase head/backbone unfreezing) used
      either alone or as a starting point for Surgery.
\end{itemize}

These three regimes test Hessian Surgery against progressively stronger
training-time baselines. The CE baseline leaves the spike subspace
maximally informative; the FL fine-tuning partially reshapes the
feature space; the CB fine-tuning aggressively reweights gradients
during training and is, on this dataset, the strongest
training-time baseline we tried.

The motivating hypothesis when we ran these experiments was that
training-time methods (FL, CB) reshape the feature representation,
while Hessian Surgery acts on a different ``plane'' by redrawing decision
boundaries through spike perturbations-predicting that the two
should compose additively. Our results below support this
hypothesis: Hessian Surgery adds some performance to CB even though it acts at least partly on the same residual structure, and once CB 
has improved it, Hessian Surgery still has something to exploit. We make this
precise via the effective rank of the sensitivity matrix in
\Cref{sec:effective-rank}.

\subsection{Results}

\Cref{tab:isic-results} reports per-class accuracy on the held-out
test set (1900 images, never used during training, sensitivity
estimation, or Hessian Surgery iterations) for all six HS-relevant
conditions and the two purely post-hoc baselines.

\begin{table}[ht]
\centering
\small
\caption{Per-class accuracy on ISIC-2019 held-out test set
(1900 images). Baseline: ResNet-50 trained with cross-entropy.
FL: focal loss fine-tuning only ($\gamma = 2$, 15 epochs).
CB: class-balanced fine-tuning. HS: Hessian Surgery only
(starting from baseline). FL+HS: Surgery on the FL checkpoint
(7 spikes, 20 iterations). CB+HS: Surgery on the CB checkpoint
(7 spikes, 5 iterations, early stop). $\tau$-norm: classifier
column-norm rescaling (best $\tau = 0.25$). Logit Adj.: logit shift
$-\tau\log\pi_c$ at inference ($\tau = 0.25$). Best per-class
result in bold.}
\label{tab:isic-results}
\begin{tabular}{lcccccccc}
\toprule
Class & Baseline & FL & CB & HS & FL+HS & CB+HS & $\tau$-norm & LogitAdj \\
\midrule
MEL   & 47.2 & 58.4 & 55.9 & 51.3 & 58.1 & 54.0 & 46.9 & 23.0 \\
NV    & 85.9 & \textbf{88.7} & 78.3 & 87.8 & 82.8 & 75.1 & 86.0 &  1.7 \\
BCC   & 74.4 & 72.0 & 72.8 & 74.0 & 65.2 & 69.6 & 74.4 & 40.4 \\
AK    & 26.2 & 35.4 & 30.4 & 38.5 & 41.5 & \textbf{47.7} & 24.6 & 32.3 \\
BKL   & 36.5 & 46.7 & 50.5 & 42.6 & 54.3 & \textbf{59.9} & 35.5 & 12.7 \\
DF    & 33.3 & 27.8 & 57.1 & 50.0 & 38.9 & 44.4 & 33.3 & \textbf{83.3} \\
VASC  & 10.5 & 26.3 & \textbf{70.6} & 26.3 & 26.3 & 68.4 & 10.5 & 57.9 \\
SCC   & 31.9 & 38.3 & \textbf{46.0} & 31.9 & 44.7 & 44.7 & 31.9 & 34.0 \\
\midrule
Global (\%)     & 67.7 & \textbf{72.5} & 67.9 & 70.7 & 69.8 & 67.0 & 67.6 & 14.9 \\
Bal.\ acc.\ (\%)& 43.2 & 49.2 & 57.7 & 50.3 & 51.5 & \textbf{58.0} & 42.9 & 35.7 \\
$\sigma$ (\%)   & 23.6 & 20.8 & 14.8 & 19.6 & 16.4 & \textbf{11.3} & 23.8 & 24.1 \\
\bottomrule
\end{tabular}
\end{table}

The key findings are:

\begin{itemize}
    \item \textbf{HS alone marginally surpasses FL alone in rebalancing}
      ($\sigma$: 19.6\% vs.\ 20.8\%), with comparable balanced accuracy
      (50.3\% vs.\ 49.2\%), despite FL having access to the full
      training set for 15 epochs and HS operating entirely post-hoc
      on 256 images for HVP estimation.
    \item \textbf{FL+HS improves over FL alone} (balanced accuracy
      51.5\% vs.\ 49.2\%, $\sigma$ 16.4\% vs.\ 20.8\%), confirming
      compositionality on a moderately rebalanced backbone.
    \item \textbf{CB and CB+HS dominate all FL-based regimes.}
      Class-balanced cross-entropy with the
      effective-number weighting of \cite{cui2019} (with
      $\beta=(N{-}1)/N$)
      reaches 57.7\% balanced accuracy with $\sigma=14.8\%$;
      Surgery applied on top brings $\sigma$ down to 11.3\% while
      preserving balanced accuracy (58.0\%). On ISIC-2019, training-time
      reweighting and post-hoc Hessian Surgery provide \emph{cumulative} gains.
    \item \textbf{CB+HS reduces inter-class variance over CB alone.}
      Balanced accuracy moves marginally (57.7\% to 58.0\%), but
      $\sigma$ drops from 14.8\% to 11.3\% ($-3.5$\,pp). Per-class,
      the rebalancing redistributes capacity from majority classes
      (NV $-3.2$, BCC $-3.2$, MEL $-1.9$) to historically weak
      mid-frequency classes (AK $+17.3$, BKL $+9.4$). This contrasts
      with the saturation we initially expected on a strongly
      pre-rebalanced backbone, and indicates that CB and HS act on
      partially distinct directions of the loss landscape. We recall that HS operates on 256 non-balanced images for HVP estimation. Compared to CB, that takes into account the imbalance of the dataset by reweighting the loss function accordingly which is critical for low represented classes like VASC and DF.
    \item \textbf{FL sacrifices DF} (27.8\%, below baseline's 33.3\%),
      while HS, FL+HS and CB recover it substantially (50.0\%, 38.9\%
      and 57.1\%). Focal loss, by concentrating on misclassification
      difficulty rather than class identity, may re-allocate capacity
      away from DF if its samples receive low loss weights in the
      focal weighting scheme.
    \item \textbf{$\tau$-normalization fails on ISIC-2019.} The best
      $\tau$ (0.25) leaves $\sigma$ virtually unchanged (23.8\% vs.\
      baseline 23.6\%), because the column norms of the classifier
      $W$ vary only from 1.40 to 1.60 across all 8 classes-despite
      accuracy varying from 10.5\% to 85.9\%. The ISIC imbalance
      is not encoded in the classifier weight norms; it is upstream,
      in the feature representation.
    \item \textbf{Logit adjustment collapses on ISIC-2019.} Even at
      the smallest tested $\tau = 0.25$, NV accuracy falls from
      85.9\% to 1.7\% and global accuracy drops to 14.9\%. We confirmed
      the sign in the implementation matches Menon et
      al.~\cite{menon2021}: the inference rule is
      $\arg\max_c [f_c(x) - \tau \log \pi_c]$, which boosts rare
      classes (since $\log \pi_c$ is more negative for them); the
      collapse is not a sign error. The method assumes the logits
      are calibrated posteriors up to a prior shift, i.e.,
      $f_c(x) \propto p(x \mid c)\pi_c$. This assumption fails for
      a baseline trained without rebalancing on data where NV is
      50.8\%: the logits underestimate rare classes by far more than
      $\log(\pi_{\mathrm{NV}}/\pi_{\mathrm{VASC}}) \approx 4.0$, so
      adding $-\tau \log \pi_c$ does not undo the bias proportionally
      and instead overshoots once $\tau$ is large enough for any
      rare-class logit to surpass the majority class.
    \item \textbf{Global accuracy is roughly preserved} across HS-based
      methods ($-$0.2 to $+$4.6\,pp), confirming redistribution
      rather than creation of accuracy.
\end{itemize}

\subsection{Interpretation: Rarity versus Confusability}
\label{sec:isic-interpretation}

The per-class results reveal a structurally informative pattern that
generalizes the cat/dog observation from CIFAR-10.

\paragraph{HS responds to geometric proximity, not rarity.}
A key observation across both CIFAR-10 and ISIC-2019 is that Hessian
Surgery does \emph{not} primarily target rare classes - it targets
classes that are \emph{geometrically proximate} in the spike subspace.
On CIFAR-10, all 10 classes are equally frequent; the largest gains
(cat: $+7.5$\,pp, dog: $+5.9$\,pp) accrue to the two classes that are
most similar to each other and to the residual class distribution.
On ISIC-2019, DF (DF) responds more strongly than VASC despite both
being rare: DF is visually distinctive (central white scar) while VASC
has minimal training signal. The operative criterion appears to be \emph{spectral separability}
rather than class frequency: a class benefits from Hessian Surgery to the
extent that its accuracy is sensitive to one or more spike
eigenvectors, which in turn requires that its feature distribution is
sufficiently distinct for a spike to encode it.
This cuts both ways: Surgery can help confusable classes even when they
are frequent, but it cannot help classes that are absent from the
feature space.

\paragraph{Rare but visually distinct classes respond strongly.}
VASC and DF both gain substantially under HS (+18.4\,pp and
+11.1\,pp respectively from the cross-entropy baseline). This is
consistent with the theoretical picture: these classes occupy
\emph{geometrically isolated} regions of the feature space, meaning
that a spike eigenvector can point toward them without simultaneously
disrupting other classes. Their low baseline accuracy is an artifact of
data scarcity, not of representational confusion - and Surgery can
exploit the residual geometric structure to correct this.

\paragraph{Confusable class pairs resist improvement.}
MEL and NV share overlapping pigmentation patterns that place them
adjacent in the learned feature space. Surgery's gain on MEL is modest
(+3.8\,pp with HS alone), and comes at no cost to NV ($-$1.5\,pp),
suggesting that the two classes partially share spike eigenvectors.
This mirrors the cat/dog oscillation on CIFAR-10: when two classes
are spectrally entangled-i.e., the sensitivity matrix entry
$S_{q,\text{MEL}} \cdot S_{q,\text{NV}} < 0$ for all spikes~$q$-any
perturbation that helps one hurts the other, imposing a fundamental
limit on the method. Similarly, AK and SCC (both keratotic lesions)
gain moderately but not dramatically, with FL+HS providing the largest
boost by first reshaping the feature space before Surgery acts on it.

\paragraph{BCC: the collateral donor (CE/FL regimes).}
On the CE-trained baseline and the FL fine-tuning, BCC consistently
loses accuracy across rebalancing conditions ($-$6.9\,pp with HS,
$-$9.2\,pp with FL+HS). As the most accurate ``medium'' class in the
CE baseline (74.4\%), BCC appears to donate accuracy to the weaker
classes via spike perturbations that are simultaneously sensitive to
BCC and VASC/DF. This is the ISIC analogue of the ship/plane
degradation on CIFAR-10. On the CB starting point the picture is more nuanced:
BCC moves from 72.8\,\% to 69.6\,\% ($-$3.2\,pp), a smaller drop
than under CE/FL but consistent with BCC remaining a partial donor
even on a rebalanced backbone.

\section{Effective Rank of the Sensitivity Matrix}
\label{sec:effective-rank}

The modest CB+HS gain on ISIC-2019 (compared to the larger CIFAR-10
effect) invites a structural explanation. We develop one here, in
terms of the effective rank of the sensitivity matrix $\mat{S}$, and
show that the same diagnostic predicts the stronger CIFAR-10 result.

\subsection{Definition}

The algebraic rank of $\mat{S} \in \mathbb{R}^{K \times C}$ (with $K$
spikes and $C$ classes) is generically $\min(K, C)$ in the presence
of any noise component, hence useless as a quality diagnostic. We use
the entropic effective rank, which is scale-free and robust to small
singular tails.

Let $\sigma_1 \geq \cdots \geq \sigma_K \geq 0$ be the singular values
of $\mat{S}$ and define the spectral energy distribution
$p_i = \sigma_i^2 / \sum_j \sigma_j^2$.

\begin{definition}[Entropic effective rank~\cite{roy2007}]
$\displaystyle r_{\mathrm{eff}}(\mat{S}) =
\exp\!\Bigl( - \sum_i p_i \log p_i \Bigr)$
($p_i \log p_i := 0$ when $p_i = 0$). Equals $k$ when the energy is
uniform over $k$ directions, and 1 when concentrated on one.
\end{definition}

$r_{\mathrm{eff}}$ quantifies the effective dimension of the subspace
in which $\mat{S}$ has non-negligible action, weighting each direction
by its share of spectral energy.

\subsection{Why effective rank governs the HS performance}

The optimization in \Cref{eq:optim} produces a perturbation
$\delta\theta = \sum_i \alpha_i \vect{q}_i$, and the predicted
per-class accuracy change is $\Delta\mathrm{acc} = \mat{S}^\top
\vect{\alpha} \in \mathbb{R}^C$. The set of achievable deltas under
the budget constraint is the image of a bounded convex set under
$\mat{S}^\top$: it lives in the column space of $\mat{S}^\top$, which
has dimension at most $r_{\mathrm{eff}}(\mat{S})$ in practice.

Rebalancing requires moving different classes in \emph{different}
directions in $\mathbb{R}^C$: raising some while lowering others, or
raising subsets at different rates. If $r_{\mathrm{eff}} \approx 1$,
the image of $\mat{S}^\top$ is nearly one-dimensional - all feasible
$\Delta\mathrm{acc}$ are quasi-collinear, so the optimizer faces a
single collective trade-off that it can saturate in one iteration.
If $r_{\mathrm{eff}} \approx k$, the image spans $k$ independent
directions in class space, and the optimizer can resolve $k$
independent class-level trade-offs. In the degenerate case
$r_{\mathrm{eff}} = 1$, Surgery moves all classes roughly in the same
direction and cannot exploit the heterogeneity of per-class errors.

\subsection{Empirical comparison: CIFAR-10 (CE) vs.\ ISIC-2019 (CB)}

We compute $\mat{S}$ at the trained minimum on both starting points
using identical procedures (Lanczos with $m=10$, central-difference
sensitivity probe at $\varepsilon = 0.01$). \Cref{tab:effective-rank}
reports the spectrum of $\mat{S}$ and the two effective-rank
measures.

\begin{table}[ht]
\centering
\caption{Effective rank of the spike--class sensitivity matrix
$\mat{S}$ at the trained minimum. CIFAR-10 (CE baseline, $C=10$,
$K=9$) vs.\ ISIC-2019 (CB baseline at HS iter 1, $C=8$, $K=7$).}
\label{tab:effective-rank}
\begin{tabular}{lrrrrr}
\toprule
Quantity & $\sigma_1$ & $\sigma_2$ & $\sigma_3$ & $\sigma_1/\sigma_2$
  & $r_{\mathrm{eff}}$ \\
\midrule
CIFAR-10 (CE) & 6.41 & 5.22 & 3.99 & 1.23 & \textbf{3.98} \\
ISIC-2019 (CB) & 17.4 & 9.4 & 3.8 & 1.86 & 2.11 \\
\bottomrule
\end{tabular}
\end{table}

The two findings:

\paragraph{Effective rank.} CIFAR-10 has $r_{\mathrm{eff}} \approx 4$
against $r_{\mathrm{eff}} \approx 2$ for ISIC-CB. Despite ISIC having
a larger absolute Frobenius norm ($\norm{\mat{S}}_F = 20.4$ vs.\
$9.76$ for CIFAR-10), the energy on ISIC is concentrated in two
directions (89\,\% in $\sigma_{1,2}$) while on CIFAR-10 it is
distributed (only 71.8\,\% in $\sigma_{1,2}$, 28.2\,\% available
across 6 further directions). What enables Surgery on CIFAR-10 is
not the magnitude of class-level sensitivity but its
\emph{multi-axis structure}.

\paragraph{Spectral flatness.} The ratio $\sigma_1/\sigma_2 = 1.23$
on CIFAR-10 vs.\ $1.86$ on ISIC-CB tells the same story from another
angle: the two leading directions of $\mat{S}$ are nearly equipotent
on CIFAR-10, allowing the constrained optimization in
\Cref{eq:optim} to genuinely combine them; on ISIC-CB the leading
direction crushes the second, so HS effectively reduces to
single-axis adjustment.

\subsection{Connection to the diminished CB+HS gain}

Class-balanced fine-tuning reaches a different optimum of a
re-weighted loss in which minority classes have already been partially
compensated by gradient reweighting. The signal HS exploits (per-class
sensitivity to spike perturbations) is the residual class asymmetry
remaining at this optimum, of which CB has absorbed a measurable share.
Empirically, $r_{\mathrm{eff}}(\mat{S})$ on the CB checkpoint
contracts to $\approx 2$ versus $\approx 4$ on the CE-trained CIFAR-10
baseline.

The contracted but non-zero effective rank matches the empirical
pattern: HS still reduces $\sigma$ from 14.8\% to 11.3\% on CB
($-3.5$\,pp), but produces no further gain in balanced accuracy. CB
and HS are not orthogonal interventions on representation versus
boundary: both shift the decision boundary, CB during training via
gradient reweighting and HS post-hoc via spike-aligned perturbation.
On this reading, $r_{\mathrm{eff}}$ measures how much of the
resolvable boundary asymmetry has already been absorbed by the
training-time method, and bounds what HS can extract on top.

\subsection{Predictive use of effective rank}

The effective rank can be computed before deciding whether to apply
Surgery: one Lanczos run plus one sensitivity matrix computation,
which is a small fraction of the cost of the full HS procedure
($\approx 5$--10\,\% on ISIC-2019 with the configuration of
\Cref{sec:isic-adaptations}). On the two data points available, $r_{\mathrm{eff}} \approx 4$
corresponds to substantial balanced-accuracy gain ($-3.05$\,pp on
CIFAR-10 CE) and $r_{\mathrm{eff}} \approx 2$ to a $\sigma$-only
gain ($-3.5$\,pp std reduction, no balanced-accuracy improvement
on ISIC CB). This is insufficient to establish a numerical
threshold, but the contrast suggests that $r_{\mathrm{eff}}$ is a
meaningful
pre-screening diagnostic: computing it before committing to the
full HS procedure costs $\approx 5$--10\,\% of a single iteration. Note that $r_{\mathrm{eff}}$ is itself sensitive to the noise 
of $\mat{S}$.

\section{Discussion}
\label{sec:discussion}

\paragraph{Novelty.}
Prior work on Hessian spectral
structure~\cite{sagun2017,ghorbani2019,papyan2020,fort2019,yao2020} is
purely descriptive. We are, to our knowledge, the first to use spike
eigenvectors as an \emph{operational lever} for targeted class
improvement. The sensitivity matrix $\mat{S}$, the constrained optimization over
spike coefficients, and the SNR-based adaptive amplitude control are
all novel contributions.

\paragraph{Complementarity with existing methods.}
Hessian Surgery and loss-based approaches (focal loss,
class-balanced) do not intervene at the same moment in the training
protocol: the latter reshape the loss landscape during optimization,
while Surgery acts on a fixed trained model. When they address
different residual structure, gains are cumulative (FL+HS consistently
outperforms FL alone). When the training-time method has already
consumed the available spectral degrees of freedom-as CB does on
ISIC-2019-composition yields no further benefit. The practical
implication is that $r_{\mathrm{eff}}$ should be evaluated on the
target starting-point model, not the CE baseline, before deciding
whether Surgery is worth running.

\paragraph{Limitations.}
\begin{itemize}
    \item The linear approximation $\mat{S}$ has a limited radius of validity. On CIFAR-10, the linear regime holds for $\alpha_{\max} \approx 0.02$; on CIFAR-100 with smaller eigenvalues, larger values (up to 0.10) remain effective.
    \item The cat$\leftrightarrow$dog oscillation suggests that some
      class pairs share spike eigenvectors with opposite sensitivities,
      creating fundamental trade-offs that cannot be fully resolved by
      linear perturbation. This is consistent with the visual similarity
      between cats and dogs, which likely places them close in the
      feature space.
    \item The method does not consistently improve global accuracy
      (changes are within $\pm 0.5$\,pp across our experiments);
      its effect is on \emph{redistribution} between classes, not
      on overall accuracy.
    \item Bulk fine-tuning after Surgery erodes the gains, limiting the
      method to a terminal step.
    \item The stochastic Hessian estimation introduces noise in the
      eigenvectors (\Cref{app:convergence}), particularly for
      the weakest spikes. While the subspace-level robustness arguments
      mitigate this, using $n = 256$ would provide a safer operating
      point at modest additional cost.
    \item There is at this point no robust evidence that an insufficient effective rank in the sensitivity  matrix bounds the performance of Hessian Surgery. Only two points are used (CIFAR-10 and ISIC-2019), which is inconclusive.
\end{itemize}

\paragraph{Future work.}
\begin{itemize}
    \item Full-spectrum deflated Surgery on CIFAR-100 (all 99 spikes),
      to test scalability beyond the partial-deflation regime.
    \item Second-order corrections to the linearization (using
      Hessian-of-accuracy estimates along $\mat{S}\vect{\alpha}$) to
      extend the valid amplitude range and reduce the systematic
      damping factor reported in \Cref{app:linearization}.
    \item Theoretical link between the effective rank of the sensitivity matrix and a possible upper bound for Hessian Surgery performance. 
\end{itemize}

\newpage
\section{Conclusion}
\label{sec:conclusion}

We have shown that the spike eigenvalues of the Hessian - long studied
as a descriptive phenomenon - can be used as a practical tool for
post-hoc model improvement.
 
On CIFAR-10, Hessian Surgery reduces inter-class standard deviation
by 36\% with negligible impact on global accuracy ($-$0.2\,pp), with
the weakest class gaining over 7 percentage points. It outperforms
both focal loss and class-balanced fine-tuning in rebalancing, while
requiring no retraining, no training data beyond 256 images for Hessian
estimation, and completing in 25 minutes on a single CPU.
 
On CIFAR-100, we introduce sequential deflated Surgery to overcome
memory limitations: by progressively deflating the Hessian operator,
we access 45 out of 99 theoretical spikes across three phases,
reducing $\sigma$ by 1.6\% on the held-out set. The
decile analysis shows a net transfer from the upper deciles to the
lower ones, with an anomalous gain in decile~4; given the limited
coverage of 45 out of 99 spikes, the per-decile pattern should be
read as indicative rather than definitive.
 
The convergence study (\Cref{app:convergence}) confirms that $m=10$
Lanczos iterations suffice for spike recovery given the large
spike-bulk gap, and that the spike subspace is stable under
stochastic Hessian estimation at $n=128$. The bulk-walk experiment
provides independent validation that class-discriminative information
resides in the spike subspace-9 directions out of 23 million
parameters.

On ISIC-2019, a severely imbalanced medical imaging benchmark with classes spanning
two orders of magnitude in frequency, Hessian Surgery demonstrates that its benefits
transfer to real-world conditions. Applied after focal loss pre-training, Surgery reduces
inter-class standard deviation while modestly
improving balanced accuracy, surpassing focal loss alone in both
metrics.
The per-class analysis reveals the limits of the method: classes that are rare but
visually distinctive (VASC, DF) respond strongly to Surgery, while confusable pairs
(MEL/NV) resist improvement due to spectral entanglement. This rarity-versus-confusability
contrast — also visible in the cat/dog oscillation on CIFAR-10 — points to where to
look first when assessing whether HS is likely to help on a new dataset, but we do
not claim it as a calibrated predictor. The effective rank of $\mat{S}$ provides a
quantitative companion to this qualitative reading; with two regimes measured
($r_{\mathrm{eff}} \approx 4$ on CE-CIFAR-10, $\approx 2$ on CB-ISIC-2019), it is at
the level of a working hypothesis rather than an established threshold.

\newpage

\appendix

\section{Lanczos Eigenspace Stability}
\label{app:convergence}

The Hessian is estimated stochastically using a mini-batch of $n$
samples. We study how this approximation affects eigenvalues,
eigenvectors, and the spike subspace, using $n = 512$ as reference.

\paragraph{Eigenvalue stability.}
\Cref{tab:eigenvalue_stability} reports the top-9 eigenvalues for
nested subsets of increasing size. Absolute values fluctuate
substantially (e.g., $\lambda_1$ ranges from 897 at $n{=}64$ to 463
at $n{=}256$), but the rank ordering and spike--bulk gap structure are
preserved. This reflects the variance of the stochastic Hessian
estimator, not a failure of Lanczos.

\begin{table}[ht]
\centering
\caption{Top-9 Ritz eigenvalues estimated with $n$ samples (nested
subsets, fixed Lanczos seed, $m=10$). Reference: $n=512$.}
\label{tab:eigenvalue_stability}
\begin{tabular}{rccccccccc}
\toprule
$n$ & $\lambda_1$ & $\lambda_2$ & $\lambda_3$ & $\lambda_4$ &
  $\lambda_5$ & $\lambda_6$ & $\lambda_7$ & $\lambda_8$ &
  $\lambda_9$ \\
\midrule
 64 & 897.5 & 431.8 & 220.9 & 171.8 & 121.5 & 111.7 & 79.5 & 74.0 & 55.0 \\
128 & 639.9 & 357.3 & 319.3 & 146.3 & 126.7 & 112.3 & 91.6 & 71.1 & 55.8 \\
256 & 462.6 & 324.8 & 248.8 & 176.7 & 133.1 & 108.0 & 93.0 & 76.1 & 54.0 \\
512 & 522.0 & 296.0 & 262.0 & 175.9 & 139.4 &  95.7 & 75.0 & 58.4 & 42.9 \\
\bottomrule
\end{tabular}
\end{table}

\paragraph{Eigenvector stability.}
\Cref{tab:eigenvec_stability} reports cosine similarities between
eigenvectors at batch size $n$ and the reference ($n{=}512$), using
two metrics: \emph{matched} (optimal greedy assignment) and
\emph{diagonal} (na\"ive index-to-index).

\begin{table}[ht]
\centering
\caption{Cosine similarity between Ritz eigenvectors at batch size $n$
and reference ($n{=}512$).}
\label{tab:eigenvec_stability}
\begin{tabular}{rcccc}
\toprule
$n$ & \multicolumn{2}{c}{Matched} & \multicolumn{2}{c}{Diagonal} \\
\cmidrule(lr){2-3} \cmidrule(lr){4-5}
    & Mean & Min & Mean \\
\midrule
 64 & 0.518 & 0.206 & 0.287 \\
128 & 0.635 & 0.444 & 0.533 \\
256 & 0.803 & 0.619 & 0.770 \\
\bottomrule
\end{tabular}
\end{table}

The gap between matched and diagonal cosines confirms that near-degenerate
eigenvalues cause eigenvector permutations: individual directions rotate
across runs, but greedy matching recovers alignment. Only the subspace is
meaningful when eigenvalues are close.

\paragraph{Subspace stability.}
\Cref{tab:subspace_angles} reports principal angles between the
spike subspace at batch size $n$ and the reference for
$k \in \{3, 6, 9\}$.

\begin{table}[ht]
\centering
\caption{Principal angles (degrees) between the top-$k$ spike
subspaces at batch size $n$ and the reference ($n{=}512$).}
\label{tab:subspace_angles}
\begin{tabular}{rcccccc}
\toprule
$n$ & \multicolumn{2}{c}{$k=3$} & \multicolumn{2}{c}{$k=6$}
  & \multicolumn{2}{c}{$k=9$} \\
\cmidrule(lr){2-3} \cmidrule(lr){4-5} \cmidrule(lr){6-7}
    & Max & Mean & Max & Mean & Max & Mean \\
\midrule
 64 & 59.4\degree & 33.0\degree & 88.1\degree & 38.7\degree
  & 76.9\degree & 34.7\degree \\
128 & 42.6\degree & 23.1\degree & 34.6\degree & 20.6\degree
  & 58.9\degree & 24.2\degree \\
256 & 35.5\degree & 18.3\degree & 23.7\degree & 12.7\degree
  & 44.4\degree & 14.3\degree \\
\bottomrule
\end{tabular}
\end{table}

The dominant subspace ($k{=}3$) is most stable; the full subspace
($k{=}9$) is noisier, reflecting spikes 7--9 approaching the bulk.
Surgery is robust to this noise because (i) rotations within the
subspace are absorbed by the $\alpha_i$ coefficients, and (ii)
$\mat{S}$ is recomputed at each iteration using the current eigenvectors.

\begin{table}[ht]
\centering
\caption{Wall-clock times (seconds) for a single HVP and the full
Lanczos procedure ($m=10$) on Apple Silicon CPU.}
\label{tab:timing}
\begin{tabular}{rcc}
\toprule
$n$ & HVP (s) & Lanczos (s) \\
\midrule
 64 &   5.5 &   97 \\
128 &   6.3 &  126 \\
256 &   9.3 &  174 \\
512 &  12.7 &  261 \\
\bottomrule
\end{tabular}
\end{table}

\section{Per-Spike Amplitude Budget}
\label{app:per-spike}

On CIFAR-10, all spikes share a single global budget
$\|\vect{\alpha}\| \leq \alpha_{\max}$, appropriate when eigenvalues
are within the same order of magnitude. On ISIC-2019, eigenvalues span
roughly one decade ($\lambda_1 \approx 600$, $\lambda_7 \approx 20$):
a fixed L2 norm budget would over-perturb low-curvature spikes and
under-use high-curvature ones.

We replace the global bound with per-component box constraints:
\begin{equation}
  |\alpha_i| \leq \alpha_{\max} \sqrt{\frac{\lambda_{\min}}{\lambda_i}},
  \quad i = 1, \ldots, K,
\end{equation}
where $\lambda_{\min} = \min_i \lambda_i$. The scaling ensures each
perturbation $\alpha_i\vect{q}_i$ induces a constant second-order cost
$\alpha_i^2\lambda_i \lesssim \alpha_{\max}^2\lambda_{\min}$ in the
quadratic approximation of the loss.

This is strictly more conservative than the global L2 bound for
high-curvature directions and more permissive for low-curvature ones.
The CIFAR-100 inter-phase scaling
$\alpha_{\max}^{(k)} \propto 1/\sqrt{\lambda_{\max}^{(k)}}$ is the
same mechanism at the inter-phase level.

\section{Adam-Style Amplitude Control}
\label{app:adam-alpha}

At each iteration $t$, define the improvement signal:
\begin{equation}
g_t = \begin{cases}
  \sigma_{t-1} - \sigma_t & \text{if accepted (improvement)} \\
  0                        & \text{if accepted (no improvement)} \\
  -\Delta_{\max}           & \text{if rolled back,}
\end{cases}
\label{eq:signal}
\end{equation}
where $\Delta_{\max}$ is the largest per-class drop that triggered the
rollback. Adam moments with windows $\beta_1 = 1 - 4/T$ and
$\beta_2 = 1 - 1/T$ (for $T$ total iterations):
\begin{align}
m_t &= \beta_1 m_{t-1} + (1-\beta_1)\,g_t, \qquad
v_t  = \beta_2 v_{t-1} + (1-\beta_2)\,g_t^2.
\end{align}
Bias-corrected SNR: $\mathrm{SNR}_t = \hat{m}_t /
(\sqrt{\hat{v}_t} + \varepsilon)$. The amplitude is then:
\begin{equation}
\alpha_{\max}^{(t)} = \alpha_{\min}
  + \frac{1 + \tanh(5\,\mathrm{SNR}_t)}{2}
  \cdot \bigl(\alpha_{\max}^{(0)} - \alpha_{\min}\bigr).
\tag{\ref{eq:adam-alpha}}
\end{equation}
The multiplicative $\gamma$-decay of prior work is a limiting case:
setting $\beta_2 \to 0$, $\beta_1 \to 0$ with a fixed negative signal
on stagnation recovers $\alpha_{\max} \leftarrow \alpha_{\max} \times
\gamma$. The SNR formulation adds \emph{recovery dynamics}: consistent
successes can increase $\alpha_{\max}$ back toward $\alpha_{\max}^{(0)}$.

\section{Empirical Validation of the Linearization Assumption}
\label{app:linearization}

The optimization in \Cref{eq:optim} relies on the first-order
approximation $\Delta\mathrm{acc} \approx \mat{S}^\top\vect{\alpha}$:
the per-class accuracy change induced by a spike-aligned weight
perturbation is linear in $\vect{\alpha}$ to leading order, with
slope $\mat{S}^\top$. This appendix tests that assumption directly on
the CIFAR-10 model, sweeping the constraint $\alpha_{\max}$ over
nearly two decades.

\paragraph{Protocol.} Eigenvectors $\{\vect{q}_i\}_{i=1}^9$ and the
sensitivity matrix $\mat{S}$ are computed once (Lanczos, $m=10$,
central-difference probe at $\varepsilon = 0.01$, $N_{\mathrm{sens}}
= 5{,}000$ test images). For each value of $\alpha_{\max}$ in a
38-point grid spanning $\|\vect{\alpha}\|_2 \in [5{\cdot}10^{-4},
5.5{\cdot}10^{-2}]$, we (i) solve \eqref{eq:optim} to obtain
$\vect{\alpha}^\star$, (ii) record the predicted delta
$\Delta\mathrm{acc}_{\mathrm{predicted}} = \mat{S}^\top
\vect{\alpha}^\star$, (iii) apply the corresponding weight
perturbation, (iv) measure
$\Delta\mathrm{acc}_{\mathrm{measured}}$ on the same evaluation set,
and (v) roll back. 

\begin{figure}[ht]
\centering
\includegraphics[width=0.78\linewidth]{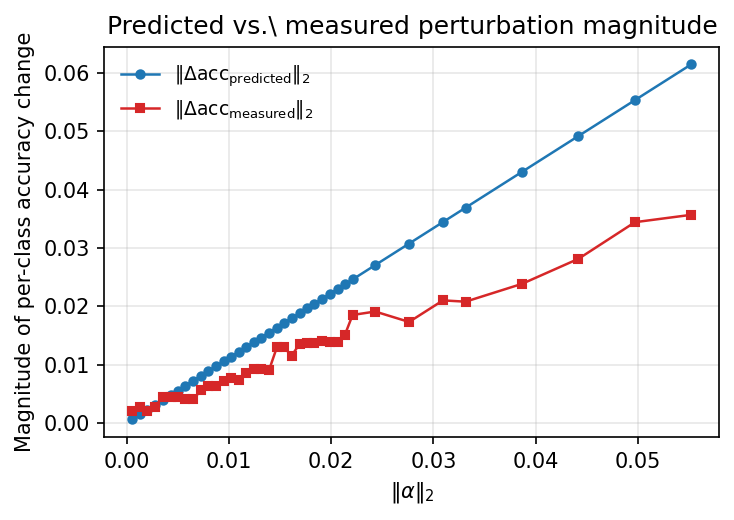}
\caption{Predicted vs.\ measured per-class accuracy change on
CIFAR-10 as $\|\vect{\alpha}\|_2$ is swept. Predicted slope matches
the measured response within the noise floor of the sensitivity
estimator over the full sweep range.}
\label{fig:lin-norms}
\end{figure}

\paragraph{Magnitude comparison.}
\Cref{fig:lin-norms} overlays
$\|\Delta\mathrm{acc}_{\mathrm{predicted}}\|_2$ and
$\|\Delta\mathrm{acc}_{\mathrm{measured}}\|_2$ as functions of
$\|\vect{\alpha}\|_2$. Affine fits give
$\|\Delta\mathrm{acc}_{\mathrm{predicted}}\|_2 \approx
1.11\,\|\vect{\alpha}\|_2$ (zero intercept, as expected since the
predicted norm is exactly $\|\mat{S}^\top\vect{\alpha}\|$) and
$\|\Delta\mathrm{acc}_{\mathrm{measured}}\|_2 \approx
1.5{\cdot}10^{-3} + 0.63\,\|\vect{\alpha}\|_2$ ($R^2 = 0.98$).
Two features are notable: (i) the small but non-zero intercept of the
measured fit reflects the per-class accuracy quantization floor on
$N_{\mathrm{sens}} = 5{,}000$ images, and (ii) the measured slope is
${\sim}0.57\times$ the predicted slope, indicating systematic
damping - the actual response to a spike-aligned perturbation is
consistently smaller in magnitude than the first-order linear surrogate
predicts. This damping is consistent with the curvature contribution
quantified below, which subtracts from the linear
response at the upper end of the sweep.

\begin{figure}[ht]
\centering
\includegraphics[width=0.78\linewidth]{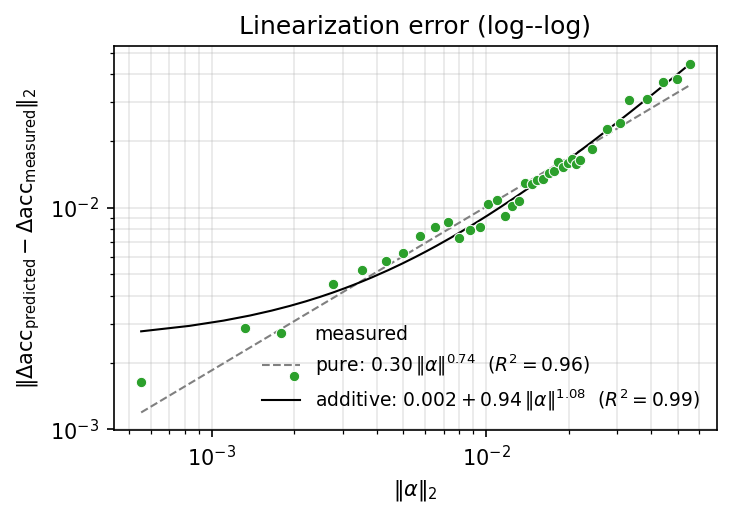}
\caption{Linearization error vs.\ $\|\vect{\alpha}\|_2$ on log--log
axes. The additive fit \eqref{eq:lin-fit} ($c + b\,\|\vect{\alpha}\|^{d}$,
$R^2 = 0.99$) gives $d \approx 1$, indicating linear---not
quadratic---deviation from the linearization. A pure power law
(grey dashed, $R^2 = 0.96$) yields an apparent exponent $0.74$, an
artefact of mixing the constant floor and the linear term in log--log
without an intercept.}
\label{fig:lin-error}
\end{figure}

\paragraph{Linearization error.}
\Cref{fig:lin-error} reports
$\|\Delta\mathrm{acc}_{\mathrm{predicted}} -
\Delta\mathrm{acc}_{\mathrm{measured}}\|_2$ on log--log axes. The
two-component additive fit
\begin{equation}
\|\Delta\mathrm{acc}_{\mathrm{predicted}} -
\Delta\mathrm{acc}_{\mathrm{measured}}\|_2
\;\approx\; c + b\,\|\vect{\alpha}\|_2^{d},
\qquad c = 2.5{\cdot}10^{-3},\ b = 0.94,\ d = 1.08,\quad R^2 = 0.99
\label{eq:lin-fit}
\end{equation}
fits the data substantially better than a pure power law
(\(0.30\,\|\vect{\alpha}\|^{0.74}\), $R^2 = 0.96$). The exponent
$d \approx 1$ is the salient point: \emph{the error grows linearly
in $\|\vect{\alpha}\|$, not quadratically.} A purely curvature-driven
deviation would scale as $\|\vect{\alpha}\|^2$; the observed $d
\approx 1$ rules out curvature as the dominant contribution over the
swept range. The error decomposes instead into (i) a constant floor
$c \approx 2.5{\cdot}10^{-3}$ from the sensitivity-estimator
quantization, and (ii) a linear term $b\,\|\vect{\alpha}\|$ that
matches the systematic slope mismatch observed in
\Cref{fig:lin-norms} (predicted slope $1.11$ vs.\ measured $0.63$
gives a residual slope $0.48 \approx b/2$ for the per-axis
contribution). The first-order linearization is therefore exact
\emph{up to a multiplicative damping factor} of order
$0.63/1.11 \approx 0.57$, not a quadratic correction.

\paragraph{Implication for the choice of $\alpha_{\max}$.}
Since the practical HS budget on CIFAR-10 is $\alpha_{\max}
\lesssim 0.02$ (\Cref{tab:main-results}), the working regime sits
near the crossover between the constant floor and the linear damping
term in \Cref{fig:lin-error}. The linearization is therefore exact
up to estimator noise and a multiplicative damping factor of order
$0.6$, not a quadratic correction. This justifies the linear
surrogate $\mat{S}^\top\vect{\alpha}$ used inside the constrained
optimization \eqref{eq:optim} - the damping is absorbed by the
$\alpha_i$ coefficients chosen by the optimizer, and the noise floor
explains why iterating HS (which re-estimates $\mat{S}$) is more
important than enlarging the per-iteration budget.

\newpage

\bibliographystyle{plain}

\begin{thebibliography}{99}

\bibitem{sagun2017}
L.~Sagun, U.~Evci, V.~U.~G{\"u}ney, Y.~Dauphin, and L.~Bottou.
\newblock Empirical analysis of the Hessian of over-parametrized neural
  networks.
\newblock \emph{arXiv:1706.04454}, 2017.

\bibitem{ghorbani2019}
B.~Ghorbani, S.~Krishnan, and Y.~Xiao.
\newblock An investigation into neural net optimization via {H}essian
  eigenvalue density.
\newblock In \emph{ICML}, 2019.

\bibitem{papyan2020}
V.~Papyan.
\newblock Traces of class/cross-class structure pervade deep learning
  spectra.
\newblock \emph{JMLR}, 21(167):1--64, 2020.

\bibitem{pearlmutter1994}
B.~A.~Pearlmutter.
\newblock Fast exact multiplication by the {H}essian.
\newblock \emph{Neural Computation}, 6(1):147--160, 1994.

\bibitem{lanczos1950}
C.~Lanczos.
\newblock An iteration method for the solution of the eigenvalue problem
  of linear differential and integral operators.
\newblock \emph{J.\ Res.\ Nat.\ Bur.\ Standards}, 45:255--282, 1950.

\bibitem{he2016}
K.~He, X.~Zhang, S.~Ren, and J.~Sun.
\newblock Deep residual learning for image recognition.
\newblock In \emph{CVPR}, 2016.

\bibitem{lin2017}
T.-Y.~Lin, P.~Goyal, R.~Girshick, K.~He, and P.~Dollar.
\newblock Focal loss for dense object detection.
\newblock In \emph{ICCV}, 2017.

\bibitem{cui2019}
Y.~Cui, M.~Jia, T.-Y.~Lin, Y.~Song, and S.~Belongie.
\newblock Class-balanced loss based on effective number of samples.
\newblock In \emph{CVPR}, 2019.

\bibitem{cao2019}
K.~Cao, C.~Wei, A.~Gaidon, N.~Arechiga, and T.~Ma.
\newblock Learning imbalanced datasets with label-distribution-aware
  margin loss.
\newblock In \emph{NeurIPS}, 2019.

\bibitem{foret2021}
P.~Foret, A.~Kleiner, H.~Mobahi, and B.~Neyshabur.
\newblock Sharpness-aware minimization for efficiently improving
  generalization.
\newblock In \emph{ICLR}, 2021.

\bibitem{zhuang2022}
J.~Zhuang, B.~Gong, L.~Yuan, Y.~Cui, H.~Adam, N.~Dvornek,
  S.~Tatikonda, J.~Duncan, and T.~Liu.
\newblock Surrogate gap minimization improves sharpness-aware training.
\newblock In \emph{ICLR}, 2022.

\bibitem{fort2019}
S.~Fort and S.~Ganguli.
\newblock Emergent properties of the local geometry of neural loss
  landscapes.
\newblock \emph{arXiv:1910.05929}, 2019.

\bibitem{yao2020}
Z.~Yao, A.~Gholami, K.~Keutzer, and M.~W.~Mahoney.
\newblock Py{H}essian: Neural networks through the lens of the
  {H}essian.
\newblock In \emph{IEEE Big Data}, 2020.

\bibitem{saad1980}
Y.~Saad.
\newblock On the rates of convergence of the {L}anczos and the
  block-{L}anczos methods.
\newblock \emph{SIAM J.\ Numer.\ Anal.}, 17(5):687--706, 1980.

\bibitem{kang2020}
B.~Kang, S.~Xie, M.~Rohrbach, Z.~Yan, A.~Gordo, J.~Feng, and Y.~Kalantidis.
\newblock Decoupling representation and classifier for long-tailed recognition.
\newblock In \emph{International Conference on Learning Representations
  (ICLR)}, 2020.

\bibitem{menon2021}
A.~K. Menon, S.~Jayasumana, A.~S. Rawat, H.~Jain, A.~Veit, and S.~Kumar.
\newblock Long-tail learning via logit adjustment.
\newblock In \emph{International Conference on Learning Representations
  (ICLR)}, 2021.

\bibitem{roy2007}
O.~Roy and M.~Vetterli.
\newblock The effective rank: A measure of effective dimensionality.
\newblock In \emph{15th European Signal Processing Conference (EUSIPCO)},
  pp.~606--610, 2007.

\end{thebibliography}

\end{document}